%% file: main.tex
\documentclass[10pt,twocolumn,letterpaper]{article}

\usepackage[pagenumbers]{cvpr} 

\usepackage{graphicx}
\usepackage{afterpage}
\usepackage{tabularx}
\usepackage{makecell}
\usepackage{amsmath} 
\usepackage[ruled]{algorithm2e}
\usepackage{adjustbox}
\usepackage{multirow}
\usepackage{pifont}

\input{preamble}

\definecolor{cvprblue}{rgb}{0.21,0.49,0.74}
\usepackage[pagebackref,breaklinks,colorlinks,allcolors=cvprblue]{hyperref}

\def\paperID{32731} 
\def\confName{CVPR}
\def\confYear{2026}

\title{Self-supervised Dynamic Heterogeneous Degradation Modeling for \\ Unified Zero-Shot Image Restoration}

\def\spaces{~~~~~~}
\author{XiaoWan~Hu\textsuperscript{1,}\thanks{Equal contribution. ~~\textsuperscript{\dag}Corresponding author.}\spaces{}Jing~Yang\textsuperscript{1,}\footnotemark[1]\spaces{}HeNan~Liu\textsuperscript{1}\spaces{}HuaQiu~Li\textsuperscript{2}\spaces{}Mai~Xu\textsuperscript{1,}\(^{\dag}\)\\
\textsuperscript{1}Beihang University\spaces{}\textsuperscript{2}Tsinghua University\\
{\tt\small \{huxiaowan, jing\_yang, lhn21373089, maixu\}@buaa.edu.cn, lihuaqiu2025@gmail.com}
}

\begin{document}
\maketitle
\input{sec/0_abstract}    
\input{sec/1_intro}

\input{sec/2_related}

\input{sec/3_observation}
\input{sec/4_method}
\input{sec/5_experiment}

\input{sec/6_Conclusion}

{
    \small
    \bibliographystyle{ieeenat_fullname}
    \bibliography{main}
}

\input{sec/X_suppl}

\end{document}

%% file: preamble.tex
\usepackage{multirow}
\usepackage{graphicx}
\usepackage{amsmath}
\usepackage{color} 
\usepackage{colortbl}
\usepackage{booktabs}
\usepackage{arydshln}
\usepackage{tabularx}
\usepackage{adjustbox}
\usepackage{makecell}
\usepackage{amssymb}
\usepackage{amsmath}
\usepackage{amsthm}

\newlength{\vspacelength}

%% file: sec/0_abstract.tex
\begin{abstract}
    Zero-shot image restoration provides a flexible way to handle diverse degradations without task-specific training. However, existing methods typically rely on stacked layers or pre-trained features to enhance degradation expression, while overlooking physically consistent priors. The insufficient degradation prompts impose the heavy training burden and high sampling costs during zero-shot diffusion. Moreover, the fixed inference trajectory often collapses to suboptimal solutions under complex corruptions. We observe that heterogeneous degradations can be reparameterized into a minimal set of physically coherent parameters for compact representation. Based on this insight, we first propose a unified physical zero-shot image restoration (UP-ZeroIR) framework that explicitly models heterogeneous degradations into a homogeneous all-in-one distribution. The distribution can be optimized directly in the latent space, enabling principled solution exploration and effective prompt adaptation. Besides, we introduce a dynamic quality-refinement strategy that adaptively adjusts the diffusion trajectory for robust globally optimal convergence. Extensive experiments demonstrate that our method achieves state-of-the-art performance across both single and mixed degradations. Our code is available at \url{https://github.com/yangjinglyy/UP-ZeroIR}.
\end{abstract}

%% file: sec/1_intro.tex
\section{Introduction}
\label{sec:intro}

\begin{figure}[t]
    \centering
    \includegraphics[trim={8mm 49mm 9mm 21mm}, clip, width=\linewidth]{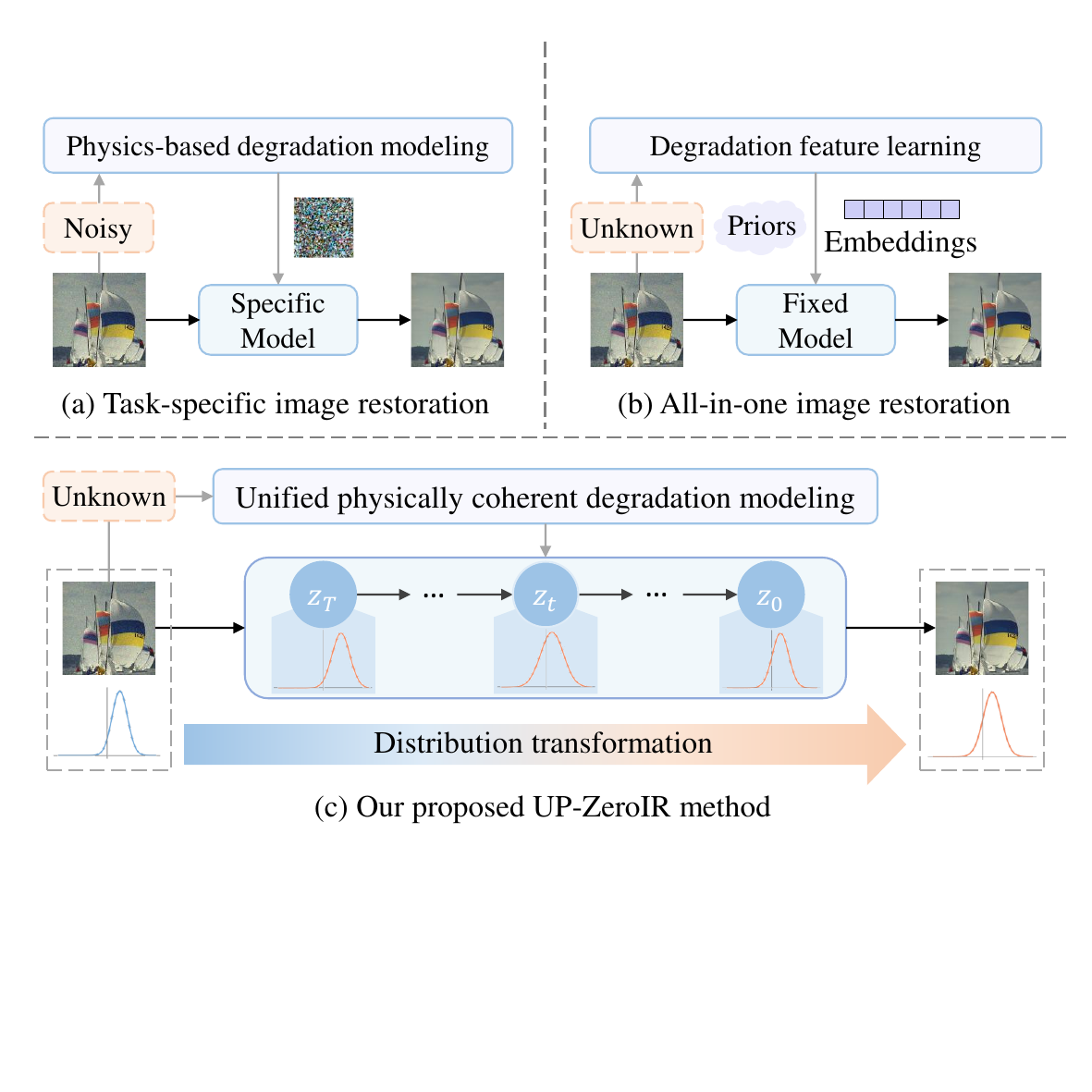}
    \vspace{-6mm}
    \caption{\textbf{Comparison of restoration pipelines.} UP-ZeroIR unifies heterogeneous degradations in latent space, steering posterior sampling toward physically consistent zero-shot restoration.}
    \label{fig:figure1}
    \vspace{-4.5mm}
\end{figure}

Image restoration is a fundamental vision task to recover clean content from degraded observations~\cite{Digitalsurvey,IRsurvey}. In the real world, images are frequently corrupted by noise, blur, haze, light, or artifacts during acquisition or transmission, which severely degrades perceptual quality~\cite{seshadrinathan2010study,tan2015video} and downstream task performance~\cite{Fu_2021_ICCV,Parmar_2022_CVPR}. Thus, effectively restoring images under diverse and unpredictable degradations remains an essential yet challenging problem.

The key challenge lies in how to handle \emph{heterogeneous} degradations in real-world scenarios without specific supervision or fixed degradation types. As shown in Fig~\ref{fig:figure1}, existing attempts generally fall into two paradigms: \textit{(i) Task-specific} models explicitly formulate degradation processes and design algorithms for a single degradation~\cite{he2010single,jiang2025msfa,li2018structure,Yang_2025_ICCV}. These methods are interpretable but inherently specialized, requiring different formulations for each degradation and failing to generalize across tasks. \textit{(ii) All-in-one} models aim to unify diverse degradations via shared architectures or learned embeddings~\cite{Li_2022_CVPR,cui2025adair,Fei_2023_CVPR,gou2024test}. However, without \emph{explicit} physical degradation modeling, they rely on deeper networks or pre-trained features to implicitly capture degradation characteristics, which constrains representation capacity and incurs high training costs as task diversity grows.

Zero-shot image restoration offers a flexible solution by removing task-specific retraining and leveraging generative diffusion priors~\cite{Fei_2023_CVPR,gou2024test,li2025ld}. However, current zero-shot models still treat degradations as black-box disturbances and rely on implicit feature learning, which leads to: \emph{(1) Weak representation}: without explicit physical prompts, the diffusion process must infer degradations from step-by-step stochastic sampling, substantially increasing the training burden when the latent space is misaligned; \emph{(2) Rigid optimization}: inference follows a fixed trajectory and prevents adaptive refinement, often collapsing into local minima and suboptimal solutions under complex corruptions. Moreover, heterogeneous degradations are often entangled, e.g., low-light conditions accompanied by noise, making feature-driven methods unable to isolate degradation factors.

Through extensive analysis of diverse real-world degradations, we observe that heterogeneous degradations, despite appearing different, can be represented by a minimal and physically coherent parameterization in the latent diffusion space. This allows us to formulate degradation modeling as a distribution-level alignment problem rather than learning high-dimensional degradation features, as shown in Fig~\ref{fig:figure1}. Estimating a unified degradation distribution is significantly more stable and efficient, and provides an explicit low-dimensional control variable that guides posterior sampling toward a physically consistent restoration process. With such an explicit representation, dynamic quality feedback during diffusion becomes feasible, enabling adaptive refinement within a unified degradation space and improving both optimization efficiency and restoration reliability.


Therefore, we propose UP-ZeroIR, a unified physical zero-shot image restoration framework that shifts from implicit feature learning to explicit degradation distribution modeling. Instead of treating heterogeneous degradations as isolated or unknown disturbances, we express diverse corruptions from the real world using a compact set of interpretable physical parameters. The transformed distribution establishes a unified solution space for all-in-one restoration. Thus, our framework can directly optimize degradation behavior in the latent diffusion space by enforcing distribution alignment, enabling controllable and degradation-aware posterior sampling rather than unguided diffusion. Furthermore, a dynamic quality-refinement strategy evaluates restoration quality during inference and adaptively adjusts the sampling trajectory to escape local minima, moving toward globally optimal solutions. Through combining explicit degradation distribution modeling with adaptive posterior refinement, UP-ZeroIR achieves physically interpretable zero-shot restoration, generalizing to unseen and mixed degradations and consistently outperforming state-of-the-art methods in both quantitative metrics and perceptual fidelity. In summary, our contributions are as follows:

\begin{itemize}\setlength{\itemsep}{3pt}
    \item We propose a unified physical parameterization of heterogeneous degradations, which enables explicit control of the latent-space distribution for zero-shot restoration.
    \item The dynamic quality-refinement strategy adaptively adjusts the diffusion trajectory in a self-assessed manner, ensuring global optimality with minimal inference costs.
    \item Extensive experiments show that UP-ZeroIR consistently outperforms prior methods on both single and mixed degradations in quantitative and perceptual performance.
\end{itemize}

%% file: sec/2_related.tex
\begin{figure}[t]
    \centering
    \begin{subfigure}{\linewidth}
        \centering
        \includegraphics[trim={2mm 8mm 2mm 5mm}, clip, width=\linewidth]{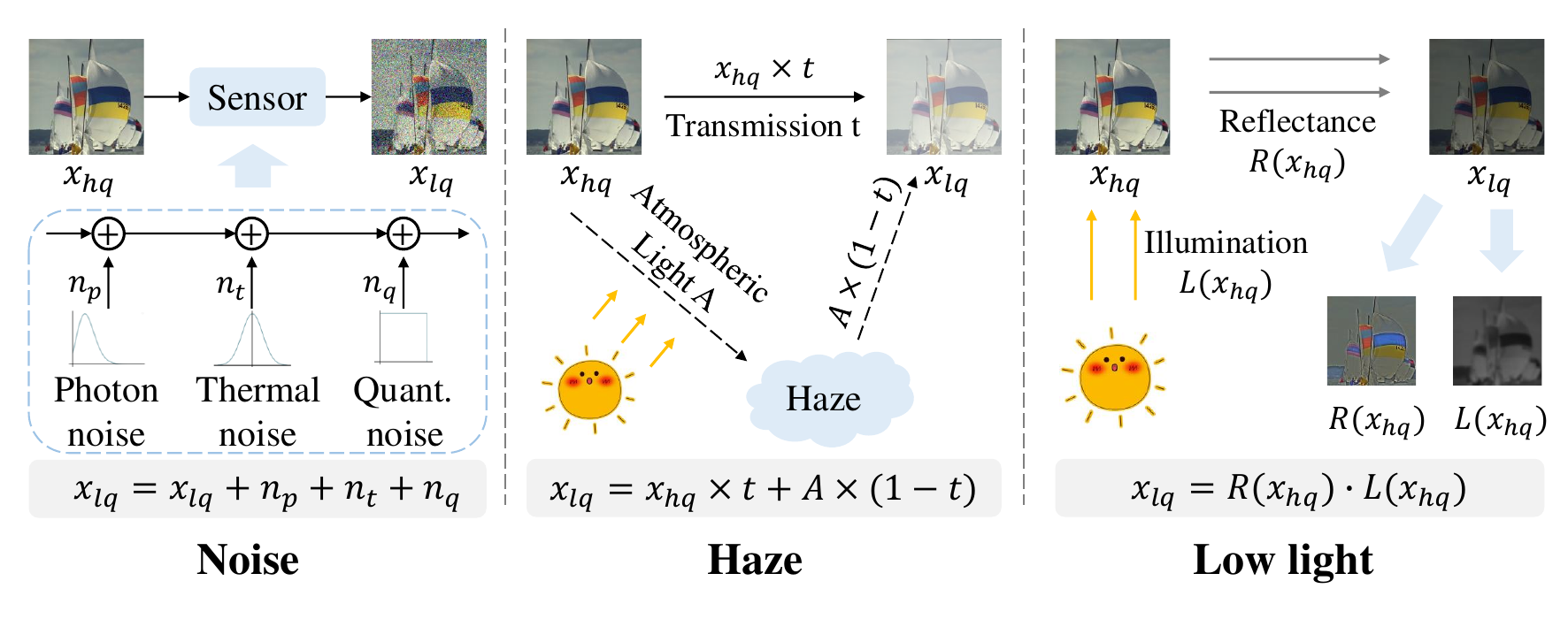}
        \caption{Toy examples illustrating different degradation transformations.}
        \label{fig:baseline_degradation_model}
    \end{subfigure}
    \begin{subfigure}{\linewidth}
        \centering
        \includegraphics[trim={1mm 1mm 1mm 1mm}, clip, width=\linewidth]{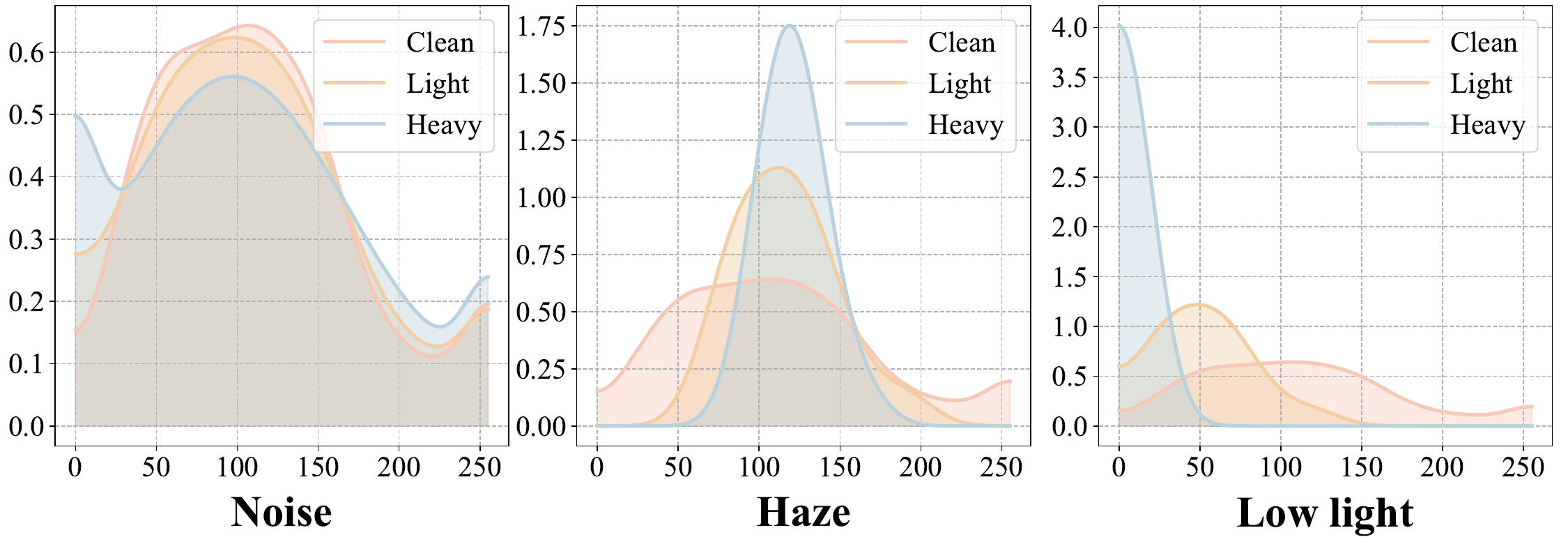}
        \caption{Distributions across various degradation types and intensities. The x-axis and y-axis represent pixel values and probability density, respectively.}
        \label{fig:baseline_degradation_distribution}
    \end{subfigure}
    \vspace{-6mm}
    \caption{\textbf{Visualization of diverse image degradations.} Each degradation type includes three intensities: clean, light, and heavy.}
    \vspace{-4mm}
\end{figure}
\begin{figure*}[t]
    \centering
    \includegraphics[trim={1mm 1mm 1mm 0mm}, clip, width=0.99\linewidth]{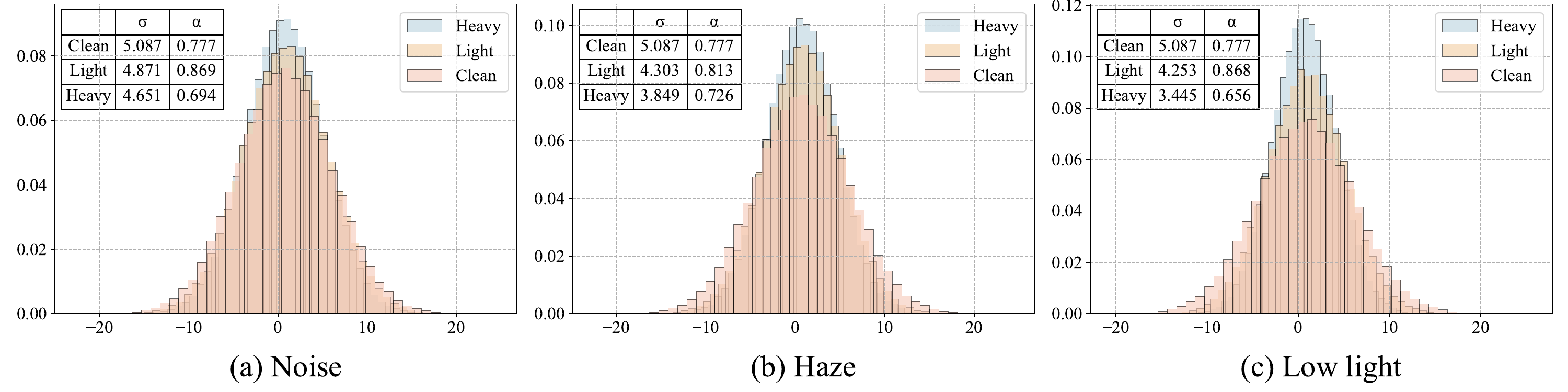}
    \vspace{-3.5mm}
    \caption{\textbf{Homogeneous degradation distribution.}
    The x-axis represents the latent representation, while the y-axis denotes the probability density. Parameters \(\sigma\) and \(\alpha\) define the dispersion and tail behavior of the distribution, respectively.}
    \label{fig:unified_degradation}
    \vspace{-4mm}
\end{figure*}

\section{Related Work}
\label{sec:related}


\noindent\textbf{Task-specific Image Restoration.}
A large body of work tackles a \emph{single} degradation with task-tailored, typically supervised architectures and objectives across denoising, deblurring, dehazing, and even super-resolution~\cite{Dabov2007BM3D,Zhang2017DnCNN,Lim2017EDSR,Chen_2023_CVPR,Liu_2024_CVPR,Zamir_2022_CVPR}. Representative advances include classical priors and residual CNNs for denoising~\cite{Dabov2007BM3D,Zhang2017DnCNN}, multi-scale and adversarial formulations that explicitly capture motion blur dynamics~\cite{Cho_2021_ICCV}, and prior-/physics-guided constraints (e.g., dark channel) with frequency-aware cues for haze removal to reduce color shifts and artifacts~\cite{he2010single,Li_2018_CVPR}. {Despite strong in-domain performance}, these pipelines presume a fixed forward model and paired supervision~\cite{xing2023daqe,chrysos2020rocgan}, transfer poorly across tasks, and remain brittle under mixed or entangled degradations, with additional task-specific tuning further limiting scalability to unified multi-task restoration.


\noindent\textbf{All-in-one Image Restoration.} 
To improve generalization across degradation types, all-in-one approaches~\cite{tang2025degradation,Li_2022_CVPR,cui2025adair,Zhang_2023_CVPR,potlapalli2023promptir,jiang2024autodir} build a unified network that shares parameters and conditions on degradation cues. Typical designs learn degradation embeddings from low-quality inputs, attach prompts or adapters to steer features, and use lightweight controllers to route pathways~\cite{Zhang_2023_CVPR,tang2025degradation,potlapalli2023promptir}. Some methods identify the degradation at test time and select submodules or statistics to match the predicted condition~\cite{cui2025adair,jiang2024autodir}. A complementary line adopts zero-shot restoration with pre-trained diffusion priors, adapting at inference without task-specific retraining by adjusting sampling schedules, guidance signals, or latent condition estimates~\cite{Fei_2023_CVPR,gou2024test,li2025ld}.  However, degradation is often modeled implicitly, control signals are not physically grounded, and inference can be sampling-heavy, limiting controllability and convergence.

\noindent\textbf{Physics-based Degradation Model.}
Recent works adopt explicit physical modeling of the image formation process to guide restoration, including noise statistics~\cite{Foi2008PG}, atmospheric scattering for haze, and illumination-reflectance decomposition rooted in Retinex theory~\cite{Fattal2008SIGGRAPH,Wei2018RetinexNet}. Representative methods instantiate these priors within learning systems, such as structure-aware formulations for dehazing~\cite{li2018structure}, extended scattering models with data-efficient training~\cite{Shyam_2023_ICCV}, and physics-based noise modeling for multispectral filter arrays~\cite{jiang2025msfa}; broader surveys synthesize such priors across tasks~\cite{pan2020physics}. Analytic constraints improve interpretability and stability. Yet reliance on task-specific forward models and precise parameters makes these methods sensitive to model mismatch~\cite{Levin2009BlindDeconv} and hard to generalize to heterogeneous or mixed degradations, so all-in-one image restoration under unseen conditions remains challenging.

%% file: sec/3_observation.tex
\section{Empirical Findings}
\label{sec:obs}

\textbf{Heterogeneous Degradation.}
Image degradation can be formulated as a linear or nonlinear transformation \( \phi \) from a high-quality image \( x_{hq} \) to its degraded observation \( x_{lq} \)~\cite{Qin_2025_CVPR}:
\begin{equation}
x_{lq} = \phi(x_{hq}) + n,
\end{equation}
where \( n \) represents stochastic noise or uncertainty.
As illustrated in \cref{fig:baseline_degradation_model}, variations in the physical degradation mechanisms result in significant discrepancies in \( \phi \).

To better understand these discrepancies, we analyze the pixel-domain distributions of three degradation types (\ie, noise, haze, and low light) on the Kodak24 dataset~\cite{franzen1999kodak}\footnote{Results for more datasets are provided in the supplementary material.}.
Specifically, high-quality images are processed using corresponding algorithms to synthesize degraded images with varying intensities.
As shown in~\cref{fig:baseline_degradation_distribution}, different degradation types and intensities exhibit distinct distributional characteristics, indicating that the inherent principles lead to unique distribution patterns in degraded images.

\begin{figure*}[ht]
    \centering
    \includegraphics[trim={29mm 22mm 31mm 27mm}, clip, width=\linewidth]{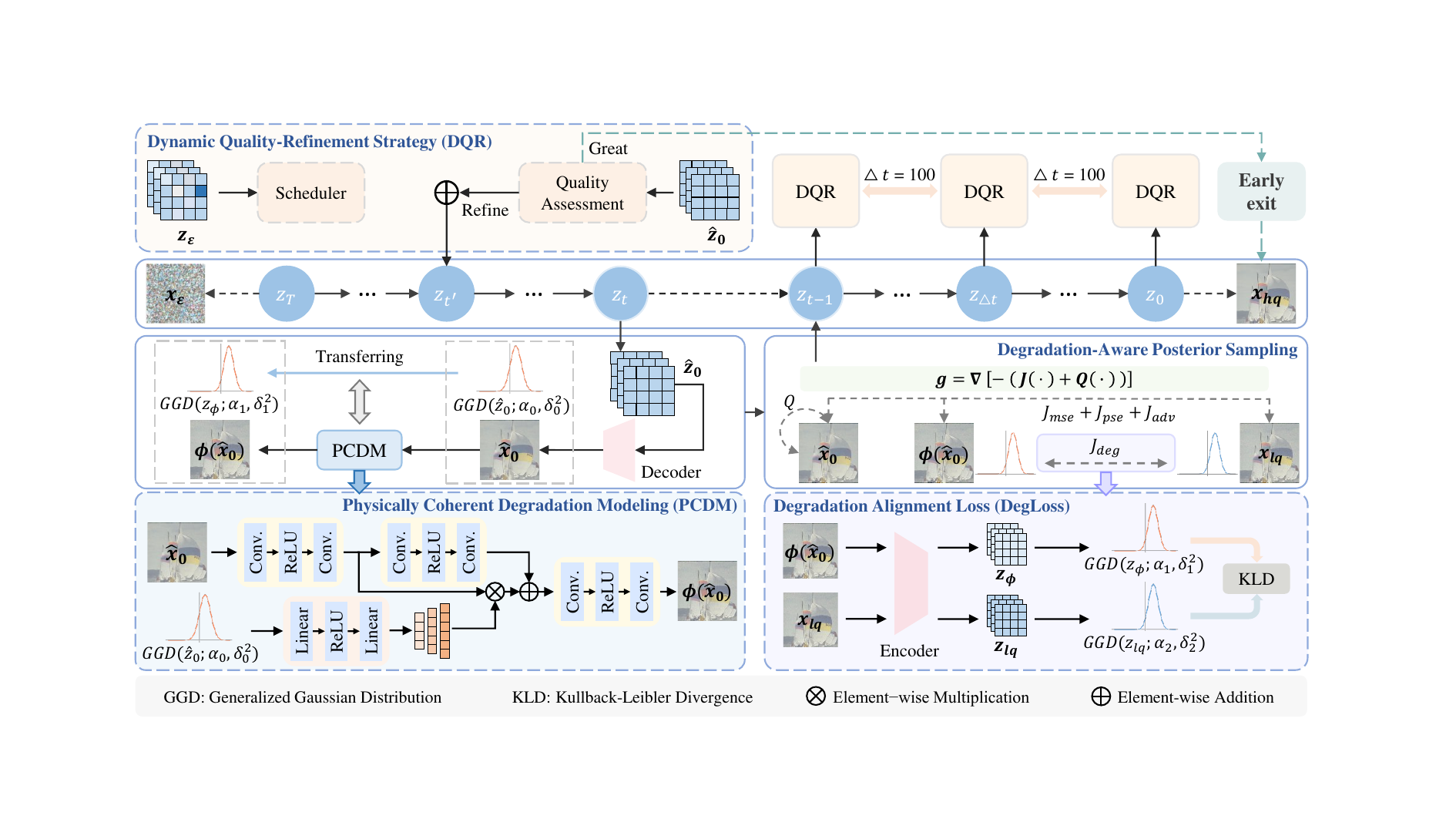}
    \vspace{-6mm}
    \caption{\textbf{Overview framework of the proposed UP-ZeroIR.} Our method takes a low-quality image \(x_{lq}\) as a conditional input to guide the correction of intermediate outputs using the pre-trained latent diffusion model, and outputs a high-quality restored image \(x_{hq}\).}
    \label{fig:framework}
    \vspace{-3mm}
\end{figure*}

\noindent\textbf{Homogeneous Translation.}
Empirically, we observe that heterogeneous degradations manifest as systematic shifts in pixel-domain distributions.
As shown in~\cref{fig:baseline_degradation_distribution}, noise results in the dispersion of the pixel distribution, haze causes a shift toward higher brightness, and low light compresses the dynamic range.
The analogous distributional behaviors indicate the potential for a unified parametric representation:
\begin{equation}
    x_{lq} \sim p(x_{lq};\psi),
\end{equation}
where $p$ is a statistical distribution with parameter set $\psi$. In practice, such commonalities amount to a homogeneous translation in distribution space across degradations. 

The pixel-domain redundancy often complicates universal modeling, while some deep networks effectively capture degradation features~\cite{wu2024ddr,liu2023evaluating,wang2020deep}.
The pre-trained latent diffusion model (LDM) offers a compact latent space that suppresses pixel-level redundancy while retaining critical cues. Thus, the invariants of degraded images can be encoded into latent diffusion space, translating pixel-domain heterogeneity into a physically coherent parametric representation.

\noindent\textbf{Physically Coherent Representation.}
We first employ the generalized Gaussian distribution (GGD) to approximate $p$ and $\psi$, which encompasses common distributions and adapts to the statistics of both natural and degraded images:
\begin{equation}
    \text{GGD}(x; \alpha, \sigma^2) = \frac{\alpha}{2\beta \Gamma(1/\alpha)} \exp \left( - \left( \frac{|x|}{\beta} \right)^{\alpha} \right),
\end{equation}
\begin{equation}
    \Gamma(z) = \int_0^\infty t^{z-1} e^{-t} \, dt \quad (z > 0), \\
\end{equation}
\begin{equation}
    \beta = \sigma \sqrt{\frac{\Gamma(1/\alpha)}{\Gamma(3/\alpha)}}, \\
\end{equation}
where \(\sigma\) and \(\alpha\) denote the scale and shape parameters.

The latent representation of degraded images is modeled with a GGD, and the parameters \(\sigma\) and \(\alpha\) are estimated on the Kodak24 dataset under the same experimental setup. As shown in~\cref{fig:unified_degradation}, latent-space distributions across degradation types and severities consistently exhibit homogeneous patterns, evidencing distribution-level alignment. The two GGD parameters serve as low-dimensional \emph{sufficient statistics} that capture the principal variation of degradations, yielding a unified and physically grounded parameterization. Consequently, we obtain a physically coherent representation that bridges heterogeneous degraded distribution and supports dynamic homogeneous modeling.



%% file: sec/4_method.tex
\section{Method}
\label{sec:method}

As shown in Fig.~\ref{fig:framework}, UP-ZeroIR is a unified physical zero-shot restoration framework, which consists of four main components: (1) the degradation-aware posterior sampling scheme runs in latent diffusion space and conditions the sampler on degradation cues; (2) physically coherent degradation modeling (PCDM) maps heterogeneous degradations to a unified low-dimensional distribution that provides the conditioning signal; (3) a distribution-alignment objective aligns the estimated latent distribution with the physical parameters, enabling controllable, degradation-aware inference; and (4) the dynamic quality-refinement (DQR) strategy adaptively adjusts the sampling trajectory using self-evaluated quality to promote globally optimal convergence. 




\subsection{Degradation-Aware Posterior Sampling}
\label{sec:method:cps}
In our framework, a pre-trained LDM is utilized to sample a Gaussian random noise map \( z_T \), and then progressively denoises it for a high-quality output in the latent space:
\begin{equation}
    q(z_{t-1} \mid z_t) = \mathcal{N}(z_{t-1}; \mu_t(z_t, \hat{z}_0), \sigma_t^2 I),
\end{equation}
\begin{equation}
    \mu(z_t, \hat{z}_0) = \frac{\sqrt{\bar{\alpha}_{t-1}} \beta_t \hat{z}_0 + \sqrt{\alpha_t} (1 - \bar{\alpha}_{t-1}) z_t}{1 - \bar{\alpha}_t},
\end{equation}
\begin{equation}
    \sigma_t^2 = \frac{1 - \bar{\alpha}_{t-1}}{1 - \bar{\alpha}_t} \beta_t, \quad \hat{z}_0 = \frac{z_t - \sqrt{1 - \bar{\alpha}_t}\,\epsilon_\theta(z_t, t)}{\sqrt{\bar{\alpha}_t}},
\end{equation}
where \( t\in \left \{ 0,...,T \right \} \), \(\bar{\alpha}_t = \prod_{i=0}^t \alpha_i, \quad \alpha_i = 1 - \beta_i\), and \(\beta_i\) denotes the magnitude of the noise introduced at the \(i\)-th timestep. The clean latent representation \( \hat{z}_0 \) is estimated from \(z_t\). \( \epsilon_\theta(z_t, t) \) indicates the predicted noise, which is the only uncertain variable in the reverse process.

To adaptively handle diverse degradations, we incorporate the input image \( x_{lq} \) to guide the direction of posterior sampling, reformulating the process as~\cite{dhariwal2021diffusion, Fei_2023_CVPR, li2025ld}:
\begin{equation}
    q(z_{t-1} | z_t, x_{lq}) \propto \mathcal{N}\left( z_{t-1} ; \mu(z_t, \hat{z}_0) + \delta g, \delta \right),
\end{equation}
where correction term \( g = \nabla_{z_t} \log p(x_{lq} \mid z_t) \) and \( \delta \) is a fixed variance. It couples the drift $\mu(z_t,\hat z_0)$ with the observation $x_{lq}$ and adapts the denoising direction to the current degradation.
\( p(x_{lq} \mid z_t) \) denotes the probability of generating the degraded observation \( x_{lq} \) given the latent representation \( z_t \), which can be approximated by \( p(x_{lq} \mid \hat{z}_0) \):
\begin{equation}
    p(x_{lq} \mid \hat{z}_0) = \frac{1}{Z} \exp\left(-\left[\lambda_1J(\phi(\hat{z}_0), x_{lq}) + \lambda_2Q(\hat{z}_0)\right]\right),
\end{equation}
where \(\phi\) denotes the transformation from high-quality to low-quality images, \(J\) denotes the image distance metric, \(Q\) evaluates the image quality, and \((Z, \lambda_1, \lambda_2) \) are scaling and weighting factors that control the guidance strength. The likelihood gradient provides degradation-aware, interpretable updates, improving robustness to mixed degradations.
As derived above, the conditional posterior sampling \( q(z_{t-1} | z_t, x_{lq}) \) can be interpreted as estimating the degradation likelihood \( p(x_{lq} \mid \hat{z}_0) \), which depends on accurately modeling the underlying degradation mechanisms.

Hence, the sampler hinges on an explicit and physically grounded likelihood, which requires sampling from a physically coherent degradation model in latent diffusion space.

\subsection{Physically Coherent Degradation Modeling}
\label{sec:method:PHDM}
Building on the analysis in \cref{sec:obs}, heterogeneous degradations admit a minimal, physically coherent parameterization in the latent diffusion space, providing a principled physical basis for modeling the complex degradation process.

As illustrated in~\cref{fig:framework}, at each sampling step \(t\), we first decode the predicted output \( \hat{z}_0 \) into the pixel domain using the LDM decoder \(D\), obtaining the corresponding image \( \hat{x}_0 = D(\hat{z}_0) \).
We then model the unified degradation distribution \( GGD(\hat{z}_0; \alpha_0, \sigma_0^2) \) of \( \hat{z}_0 \).
The distribution parameters\( (\alpha_0, \sigma_0) \) are embedded through linear layers \(l\), allowing the degradation process to be represented as a structured probabilistic distribution, thereby producing a physically plausible degraded observation \( \phi(\hat{x}_0) \):
\begin{equation}
    \phi(\hat{x}_0) = f_3(f_2(f_1(\hat{x}_0)) + l(\alpha_0, \sigma_0) \cdot f_1(\hat{x}_0)),
\end{equation}
where \( f_1, f_2, f_3 \) denote convolution layers.
We optimize the PCDM by minimizing a mixed loss function:
\begin{equation}
    \begin{aligned}
        L_{\text{total}} = & L_{\text{mse}}(\phi(\hat{x}_0), \hat{x}_0) + L_{\text{pse}}(\phi(\hat{x}_0), \hat{x}_0) \\
        & + L_{\text{adv}}(\phi(\hat{x}_0), x_{lq}),
    \end{aligned}
\end{equation}
here \( L_{\text{mse}} \), \( L_{\text{pse}} \) and \( L_{\text{adv}} \) denote the mean squared error loss, perceptual loss~\cite{johnson2016perceptual}, and adversarial loss~\cite{goodfellow2020generative}, respectively.

In this way, PCDM establishes an equivalent distribution mapping between the immediate result \( \hat{x}_0 \) and its degraded counterpart \( \phi(\hat{x}_0) \), thereby reformulating degradation modeling as a distribution-level approximation problem.

\subsection{Degradation Alignment Loss}
\label{sec:method:dal}
During posterior sampling, we develop a novel degradation alignment loss (DegLoss) through the image discrepancy term \( J(\phi(\hat{z}_0), x_{lq}) \). It casts the unified degradation distribution as an explicit and tractable optimization objective.

Given the degraded output \( \phi(\hat{x}_0) \) from the PCDM and the input \( x_{lq} \), we first employ the LDM encoder \(E\) to generate their latent representations \(z_{\phi}\) and \(z_{lq}\). Secondly, we model their parametric degradation distributions as \( GGD(z_{\phi}; \alpha_1, \sigma_1^2) \)and \( GGD(z_{lq}; \alpha_2, \sigma_2^2) \).
The degradation alignment loss is defined as the Kullback-Leibler divergence (KLD) between these two distributions~\cite{liu2023evaluating, cao2024grids}:
\begin{equation}
    \begin{aligned}
        J_{deg} &= \ln \left( \frac{\alpha_1 \sigma_2 \Gamma(1/\alpha_2) \sqrt{\Gamma(1/\alpha_2) \Gamma(3/\alpha_1)}}{\alpha_2 \sigma_1 \Gamma(1/\alpha_1) \sqrt{\Gamma(1/\alpha_1) \Gamma(3/\alpha_2)}} \right) \\
        &\quad + \left( \frac{\sigma_1 \sqrt{\Gamma(1/\alpha_1) \Gamma(3/\alpha_2)}}{\sigma_2 \sqrt{\Gamma(1/\alpha_2) \Gamma(3/\alpha_1)}} \right)^{\alpha_2} \frac{\Gamma(\alpha_2/\alpha_1 + 1/\alpha_1)}{\Gamma(1/\alpha_1)}.
    \end{aligned}
\end{equation}

Besides, to improve reconstruction fidelity, we combine the degradation alignment loss with mean squared error \(J_{mse}\), perceptual loss \(J_{pse}\), and adversarial loss \(J_{adv}\):
\begin{equation}
    \begin{aligned}
        J_{total} &= \lambda_1J_{deg}(\phi(\hat{x}_0), x_{lq}) + \lambda_2J_{mse}(\phi(\hat{x}_0), x_{lq}) \\
        &+ \lambda_3J_{pse}(\phi(\hat{x}_0), x_{lq}) + \lambda_4J_{adv}(\phi(\hat{x}_0), x_{lq}, \hat{x}_0),
    \end{aligned}
\end{equation}
where \( \lambda_1, \lambda_2, \lambda_3, \lambda_4 \) control the contributions of each term.
Furthermore, we future define the luminance and chrominance quality of \( \hat{x}_0 \) as the image quality term \( Q(\hat{z}_0) \):
\begin{equation}
    Q_{total} = \lambda_5 \sum_{i=1}^{3} \left( \left| \bar{x}^i_{0} - \tau \right|^2 \right) + \lambda_6 \sum_{(p,q) \in \mathcal{D}} \left( \delta(p,q) \right),
\end{equation}
where \( \bar{x}^i_{0} \) denotes the mean luminance of channel \(i\), \( \tau \) is the natural exposure standard, \( \mathcal{D} \) represents pixel pairs in the chrominance channels, \( \delta(p,q) \) measures their color difference, and \( \lambda_5, \lambda_6\) are weighting factors.
With these designs, our method effectively recasts posterior sampling toward physically consistent zero-shot restoration, enabling degradation-aware and principled solution exploration.

\subsection{Dynamic Quality-Refinement Strategy}
\label{sec:method:DQR}

To mitigate suboptimal convergence under complex corruptions during static diffusion inference, we integrate the no-reference image quality assessment model Arniqa~\cite{agnolucci2024arniqa}, pre-trained on diverse degradations from the real world, to dynamically modulate the posterior sampling strategy.

Following the evaluation paradigm presented in Fig.~\ref{fig:framework}, at each step \(j\) (performed every \( \bigtriangleup t = 100 \) steps), Arniqa is applied to compute the quality score \(s_j\) of the restored image \( \hat{x}_0 \) at the current diffusion step \(t\).
We then compare it with the score from the previous iteration \(s_{i-1}\), forming the basis for our adaptive decision-making strategy:
\begin{equation}
    \text{Decision} = 
    \begin{cases} 
        \text{Great}, & s_j - s_{j-1} < \eta \\ 
        \text{Refine}, & s_j - s_{j-1} \geq \eta,
    \end{cases}
\end{equation}
where \( \eta \) is a predefined threshold controlling the refinement sensitivity.
If the score difference falls below the threshold, the restoration quality is considered satisfactory, the inference terminates and \( \hat{x}_0 \) is regarded as the optimal output.
Otherwise, further refinement is performed adaptively: for \( t > 0 \), inference continues following standard diffusion updates; for \( t = 0 \), noise is introduced into the current latent representation \( z_0 \) to re-initiate the posterior sampling:
\begin{equation}
    z_{t'} = z_t \cdot \sqrt{a_{t} }  + \sqrt{1 - a_{t} } \cdot z_{\epsilon}, \quad z_{\epsilon} \sim \mathcal{N}(0, I),
\end{equation}
where \( a_{t} \) denotes the noise scale, and noise \( z_{\epsilon} \) is independently sampled from a standard Gaussian distribution.

\input{tab/lowlightE}
\begin{figure*}[t]
    \centering
    \includegraphics[trim={2mm 13mm 2mm 10mm}, clip, width=\linewidth]{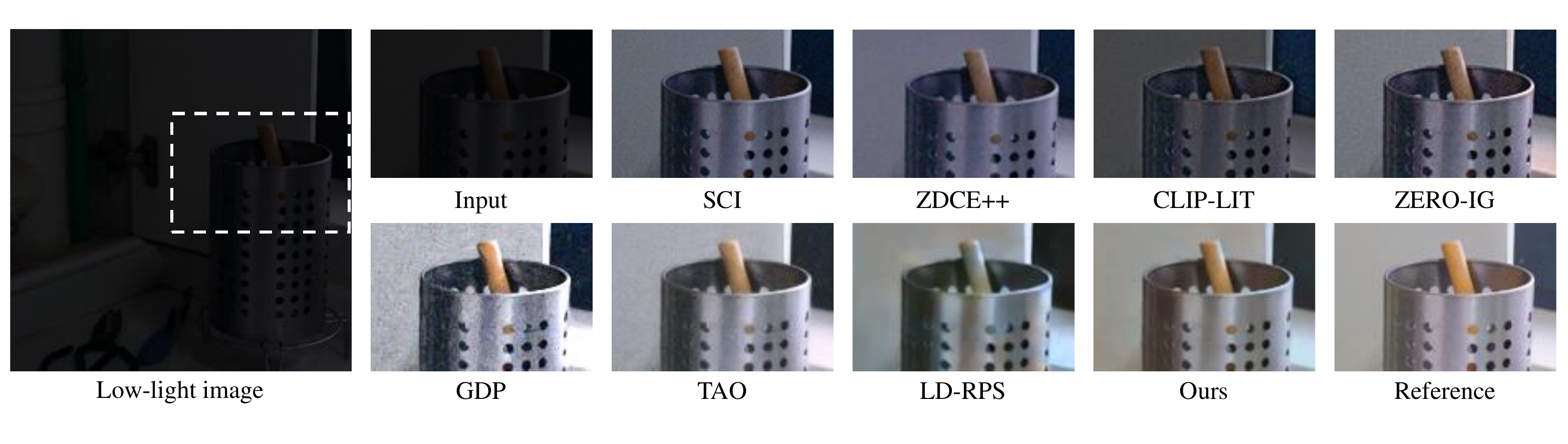}
    \vspace{-5mm}
    \caption{\textbf{Qualitative performance of low-light enhancement on the LOLv1 dataset.} Our method achieves a more natural lighting.}
    \label{fig:result_lowlight}
    \vspace{-2mm}
\end{figure*}

%% file: tab/lowlightE.tex
\begin{table*}[htbp]
\centering
\begin{adjustbox}{max width=\textwidth}
\begin{tabular}{l|ccc|ccccc|ccccc}
\noalign{\hrule height 1pt}
\multirow{2}{*}{Methods} & \multicolumn{3}{c|}{Property{\rule{0pt}{1.1em}}} & \multicolumn{5}{c|}{LOLv1} & \multicolumn{5}{c}{LOLv2} \\
\cline{2-14} 
& B{\rule{0pt}{1.1em}} & U & Z & PSNR$\uparrow$ & SSIM$\uparrow$ & LPIPS$\downarrow$ & PI$\downarrow$ & NIQE$\downarrow$ & PSNR$\uparrow$ & SSIM$\uparrow$ & LPIPS$\downarrow$ & PI$\downarrow$ & NIQE$\downarrow$ \\
\noalign{\hrule height 1pt}
\multicolumn{14}{c}{Task-specific} \\
\hline
SCI~\cite{Ma_2022_CVPR} & \ding{55} & \ding{51} & \ding{55} & 14.86 & 0.704  & 0.219 & 4.42 & 5.94  & 17.17 & 0.639 & 0.264 & 6.27 & 10.47 \\
ZDCE++~\cite{li2021learning} & \ding{55} & \ding{51} & \ding{55} & 14.38 & 0.523 & 0.240 & \textbf{4.04} & \textbf{4.95} & 16.76 & 0.428 & 0.284 & \textbf{4.17} & \textbf{5.99} \\
CLIP-LIT~\cite{Liang_2023_ICCV} & \ding{55} & \ding{51} & \ding{55} & 12.63 & 0.678 & 0.240 & 4.28 & 5.96 & 15.41 & 0.650  & 0.315 & 6.57 & 11.31 \\
ZERO-IG~\cite{Shi_2024_CVPR} & \ding{55} & \ding{51} & \ding{51} & \textbf{17.22} & \textbf{0.794} & \textbf{0.184} & 4.92 & 6.22 & \textbf{18.63} & \textbf{0.751} & \textbf{0.231} & 5.64 & 8.59 \\
\hline
\multicolumn{14}{c}{Posterior sampling} \\
\hline
GDP~\cite{Fei_2023_CVPR} & \ding{55} & \ding{51} & \ding{51} & 16.86 & 0.689 & 0.299 & 4.67 & 5.91 & 15.32 & 0.597 & 0.337 & 5.13 & 8.18 \\
TAO~\cite{gou2024test} & \ding{51} & \ding{51} & \ding{51} & 17.42 & 0.792 & 0.320 & 6.03 & 7.74 & 16.78 & 0.747 & \textbf{0.314} & 6.41 & 9.44 \\
LD-RPS~\cite{li2025ld} & \ding{51} & \ding{51} & \ding{51} & 17.26 & 0.797 & 0.291 & 5.02 & 5.79 & 18.22 & 0.744 & 0.335 & 5.05 & 6.03 \\
Ours & \ding{51} & \ding{51} & \ding{51} & \textbf{18.21} & \textbf{0.823} & \textbf{0.241} & \textbf{4.65} & \textbf{5.47} & \textbf{19.20} & \textbf{0.761} & 0.332 & \textbf{4.98} & \textbf{5.72} \\
\noalign{\hrule height 1pt}
\end{tabular}
\end{adjustbox}
\vspace{-0.5mm}
\caption{\textbf{Quantitative results of low-light enhancement on the LOLv1 and LOLv2 datasets.} The best results within each category are boldfaced. Methods are grouped by three properties: B (task-blind), U (unsupervised), and Z (zero-shot), reflecting their adaptability.}
\label{tab:lowlightE}
\vspace{-0.5mm}
\end{table*}

%% file: sec/5_experiment.tex
\section{Experiments}
\label{sec:exp}

\subsection{Implementation Details}
\label{sec:exp:details}
\begin{figure*}[t]
    \centering
    \includegraphics[trim={12mm 6mm 15mm 16mm}, clip, width=\linewidth]{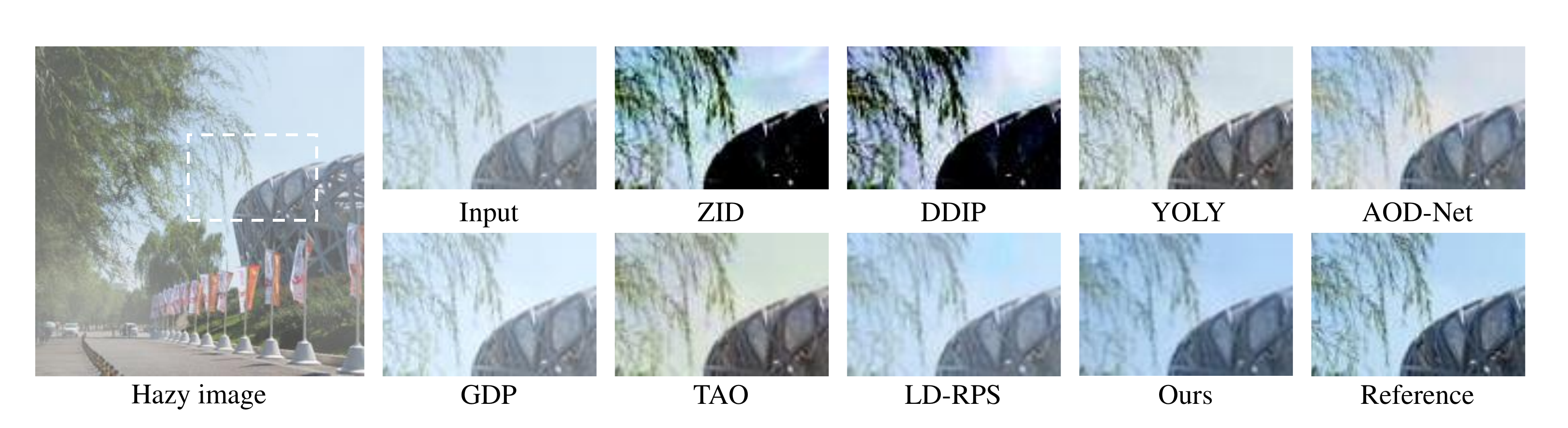}
    \vspace{-6mm}
    \caption{\textbf{Qualitative performance of image dehazing on the HSTS dataset.} Our method shows higher color fidelity and naturalness.}
    \label{fig:result_dehazing}
    \vspace{-2.0mm}
\end{figure*}
\begin{figure*}[t]
    \centering
    \includegraphics[trim={12mm 6mm 15mm 16mm}, clip, width=\linewidth]{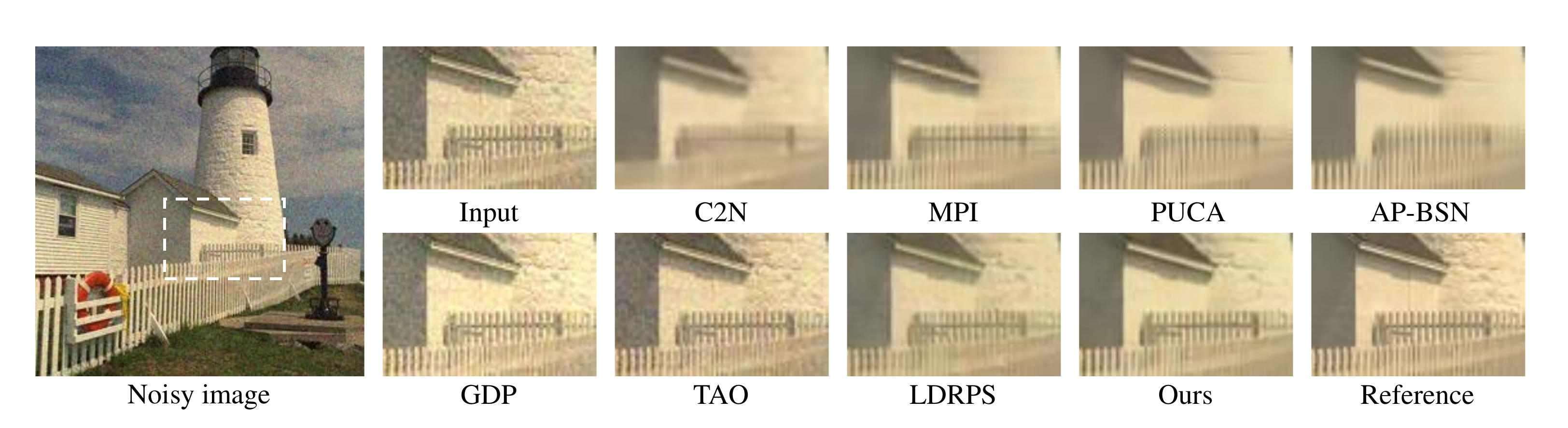}
    \vspace{-6mm}
    \caption{\textbf{Qualitative performance of image denoising on the Kodak24 dataset.} Our method yields cleaner images with better texture.}
    \label{fig:result_denoising}
    \vspace{-3.5mm}
\end{figure*}

For a comprehensive and fair evaluation, we conduct experiments on three representative image restoration tasks: denoising, dehazing, and low-light enhancement, using the same settings.
And our method is compared with widely-adopted zero-shot posterior sampling methods, including GDP~\cite{Fei_2023_CVPR}, TAO~\cite{gou2024test}, and LD-RPS~\cite{li2025ld}, as well as supervised all-in-one methods~\cite{Li_2022_CVPR, potlapalli2023promptir, Zheng_2024_CVPR}.
Following previous studies, we report PSNR, SSIM~\cite{wang2004image}, and LPIPS~\cite{zhang2018unreasonable}, along with PI~\cite{Wang_2018_ECCV_Workshops} and NIQE~\cite{mittal2012making} to assess perceptual quality in low-light enhancement.
In experiments, we employ the stable diffusion model~\cite{Rombach_2022_CVPR} with 1000 timesteps.
PCDM is optimized by Adam optimizer with a learning rate of \(1 \times 10^{-5}\) during testing.
The scaling factor \(Z\) and threshold \( \eta \) are set to 4000 and 0.01, respectively. Loss terms are reweighted via an empirically designed scheme, with details in the supplementary. All experiments are implemented through the PyTorch framework~\cite{NEURIPS2019_bdbca288} on a single NVIDIA L20 GPU.
\subsection{Experimental Results}
\label{sec:exp:results}

\textbf{Low-Light Enhancement.}
We conduct experiments on both LOLv1 and LOLv2 datasets~\cite{Wei2018RetinexNet}, comparing against task-specific methods such as SCI~\cite{Ma_2022_CVPR}, ZDCE++~\cite{li2021learning}, CLIP-LIT~\cite{Liang_2023_ICCV}, and Zero-IG~\cite{Shi_2024_CVPR}. The quantitative results are presented in \cref{tab:lowlightE}.
Our method consistently outperforms both posterior sampling and task-specific methods, achieving the highest performance on PSNR and SSIM while maintaining competitive results on other metrics. Notably, our method surpasses the second-best method, LD-RPS, by 0.95 dB and 0.98 dB in PSNR on the LOLv1 and LOLv2 datasets. Some results are comparable to supervised all-in-one methods seen in the supplementary.
Qualitative comparisons in \cref{fig:result_lowlight} show that our method effectively enhances low-light images, improving brightness and contrast while preserving natural colors and details. The over-saturation, halos, and banding in bright regions are avoided.


\input{tab/dehazing}
\noindent\textbf{Image Dehazing.}
We evaluate our method on the HSTS subset of the RESIDE dataset~\cite{li2018benchmarking}, comparing against additional task-specific methods including ZID~\cite{li2020zero}, DDIP~\cite{Gandelsman_2019_CVPR}, YOLY~\cite{Li:2021kt}, and AOD-Net~\cite{Li_2017_ICCV}.
The quantitative results are reported in \cref{tab:dehazing}.
Our method outperforms all zero-shot and unsupervised baselines, improving PSNR by 1.03 dB over the second-best method, LD-RPS, with similar gains on other metrics.
Due to the absence of relevant degradation priors, GDP fails to remove haze. Moreover, we present qualitative comparisons in \cref{fig:result_dehazing}.
Compared with other specific, all-in-one, or even supervised methods, our method effectively removes haze while preserving fine image details, restores distant depth and faithful sky gradients, and corrects color shifts without over-dehazing. It also avoids typical artifacts, yielding clearer and natural-looking results.


\input{tab/denoising}
\input{tab/ablation}
\noindent\textbf{Image Denoising.}
We incorporate C2N~\cite{Jang_2021_ICCV}, MPI~\cite{ma2024masked}, PUCA~\cite{jang2023puca}, and AP-BSN~\cite{Lee_2022_CVPR} as task-specific baselines, using the Kodak24 dataset for testing.
As presented in \cref{tab:denoising}, our method consistently outperforms both posterior sampling and task-specific methods across all metrics, and delivers performance comparable to some supervised all-in-one methods.
Our method achieves a 0.85 dB improvement in PSNR over the second-best method, LD-RPS.
In contrast, GDP exhibits limited denoising capability due to its inadequate noise modeling.
Qualitative results in \cref{fig:result_denoising} illustrate that our method effectively suppresses the coarse noise on the wall while preserving fine structural details, such as the fence patterns, wall shading transitions, and edge contours, thereby avoiding the detail loss caused by over-smoothing.


In summary, our physically coherent alignment can suppress model collapse and enable stable restoration across diverse conditions. This unified distribution model achieves superior performance while preserving textures, colors, and visual fidelity under realistic heterogeneous degradations.


\input{tab/mixed}

\subsection{Ablation Study}
\noindent\textbf{Component Ablation.}
In this section, we evaluate the effectiveness of three key components in our method and design the following ablation experiments: \emph{(1) w/o PCDM}: we replace the physically coherent degradation model with ResBlocks of similar parameter count; \emph{(2) w/o DegLoss}: we replace the degradation alignment loss with a standard pixel-level L1 loss; \emph{(3) w/o DQR}: we remove the dynamic quality-refinement strategy and employ a fixed sampling schedule with 1000 timesteps.
All models are evaluated using the same experimental settings for fair comparison.

As presented in \cref{tab:ablation}, replacing PCDM with parameter-matched ResBlocks causes a marked drop, indicating the importance of incorporating explicit physical priors, as empirically analyzed in~\cref{sec:obs}.
Similarly, substituting DegLoss with a pixel-wise L1 loss weakens distribution-level alignment and reduces accuracy, underscoring its role in physically interpretable guidance.
Finally, removing DQR and using a fixed schedule yields suboptimal convergence, validating its benefit for quality-driven global optimization. Overall, the three components provide complementary gains, and ablating any one leads to clear regressions.

\begin{figure}[t]
    \centering
    \includegraphics[trim={0mm 0mm 0mm 0mm}, clip, width=\linewidth]{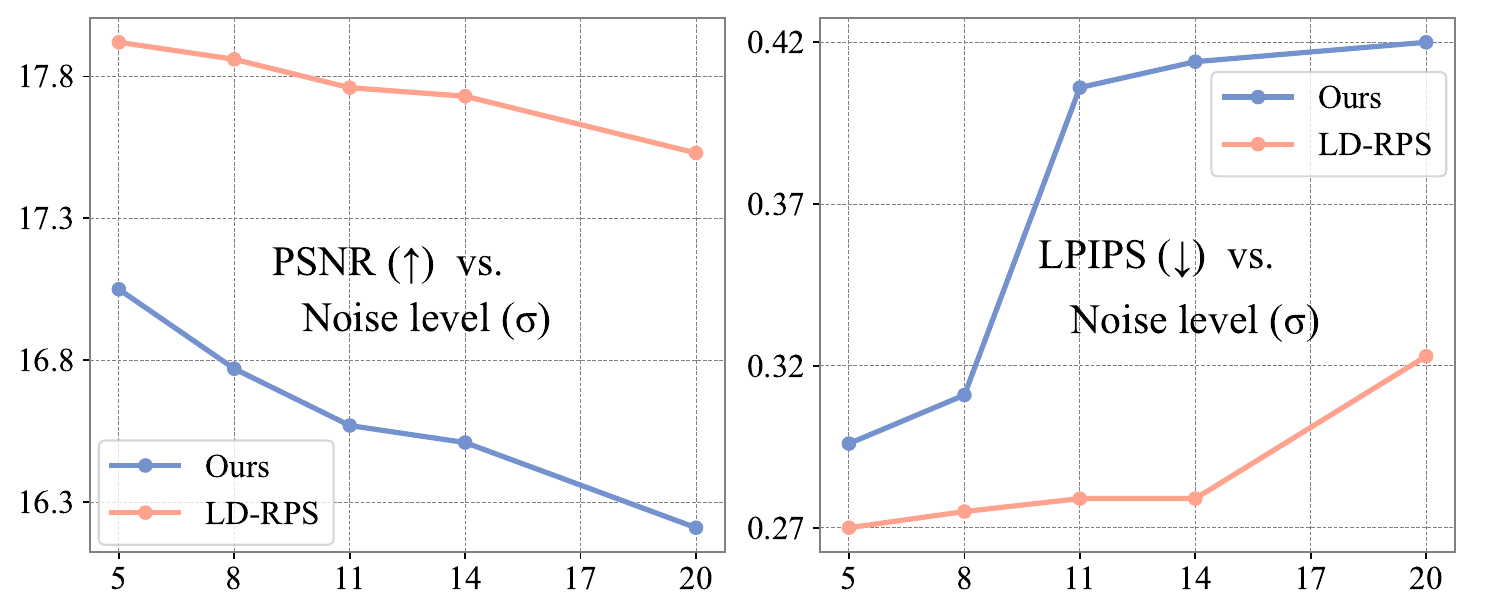}
    \vspace{-5mm}
    \caption{\textbf{Noise robustness under mixed degradations.} Noise level is defined by the standard deviation (\(\sigma\)) of Gaussian noise.}
    \vspace{-4mm}
    \label{fig:noise_robustness}
\end{figure}

\noindent\textbf{Mixed Degradation Robustness.}
Real-world images always routinely mix degradations, so we construct mixed-degradation datasets based on LOLv1 by incorporating haze and noise into low-light images, thereby generating two complex degradation scenarios: Low-light + Noise and Low-light + Haze + Noise.
We compare our method with other posterior sampling methods and present the results in~\cref{tab:mixed}.
Our method consistently achieves superior performance, while other posterior sampling methods suffer from performance drops when handling entangled degradations.
Furthermore, we introduce noise perturbation on the Low-light + Haze + Noise scenario, as illustrated in~\cref{fig:noise_robustness}. As the noise intensity increases, our method maintains more stable performance compared to the second-best method, LD-RPS, whose performance deteriorates notably.
In summary, these findings highlight strong robustness to mixed degradations and noise perturbations, supporting the effectiveness of our physically coherent degradation modeling.

\begin{figure}[t]
    \centering
    \includegraphics[trim={16mm 18mm 15mm 8mm}, clip, width=\linewidth]{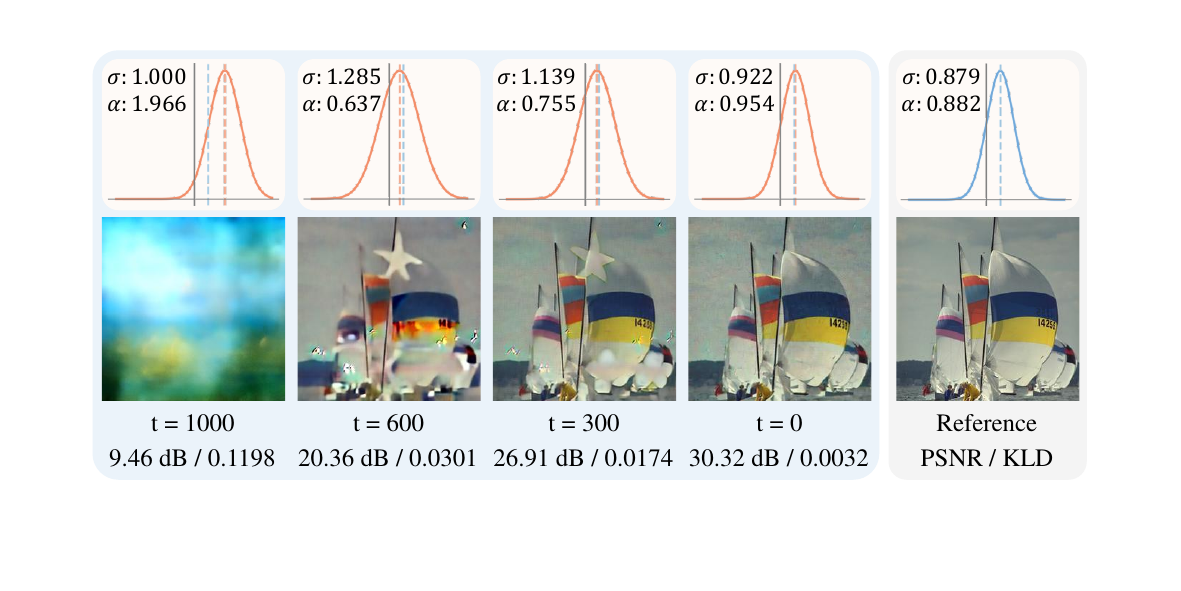}
    \vspace{-5mm}
    \caption{\textbf{Intermediate visualization of the denoising process.} t is the latent diffusion time step. We report PSNR and KLD as measures of image- and distribution-level distance to the reference.}
    \vspace{-4mm}
    \label{fig:visual}
\end{figure}

\noindent\textbf{Step-wise Distribution Visualization.}
We visualize the intermediate results during the reverse denoising process, as illustrated in \cref{fig:visual}.
As the diffusion process evolves from Gaussian noise to a high-quality image, the unified degradation distribution progressively converges along a stable and physically consistent trajectory toward the clean degradation distribution.
Our method directly optimizes degradation behaviors within the latent space through explicit distribution alignment, enabling controllable and degradation-aware solution exploration during posterior sampling.
As a result, our method effectively handles various heterogeneous and mixed degradations, alleviates training overhead, and generates high-fidelity, physically plausible outputs.

%% file: tab/dehazing.tex
\begin{table}[tb]\small
\centering
\begin{adjustbox}{max width=\linewidth}
\begin{tabular}{l|ccc|ccc}
\noalign{\hrule height 1pt}
\multirow{2}{*}{Methods} & \multicolumn{3}{c|}{Property{\rule{0pt}{1.1em}}} & \multicolumn{3}{c}{HSTS} \\
\cline{2-7}
 & B{\rule{0pt}{1.1em}} & U & Z & PSNR$\uparrow$ & SSIM$\uparrow$ & LPIPS$\downarrow$ \\
\noalign{\hrule height 1pt}
\multicolumn{7}{c}{Task-specific} \\
\hline
ZID~\cite{li2020zero} & \ding{55} & \ding{51} & \ding{51} & 19.31 & 0.796 & 0.191 \\
DDIP~\cite{Gandelsman_2019_CVPR} & \ding{55} & \ding{51} & \ding{51} & 20.20 & 0.846 & 0.150 \\
YOLY~\cite{Li:2021kt} & \ding{55} & \ding{51} & \ding{51} & \textbf{20.49} & 0.794 & \textbf{0.108} \\
AOD-Net~\cite{Li_2017_ICCV} & \ding{55} & \ding{55} & \ding{55} & 19.15 & \textbf{0.860} & 0.129 \\
\hline
\multicolumn{7}{c}{Posterior sampling} \\
\hline
GDP~\cite{Fei_2023_CVPR} & \ding{55} & \ding{51} & \ding{51} & 12.57 & 0.703 & 0.164 \\
TAO~\cite{gou2024test} & \ding{51} & \ding{51} & \ding{51} & 15.66 & 0.775 & 0.216 \\
LD-RPS~\cite{li2025ld} & \ding{51} & \ding{51} & \ding{51} & 20.48 & 0.804 & 0.173 \\
Ours & \ding{51} & \ding{51} & \ding{51} & \textbf{21.51} & \textbf{0.820} & \textbf{0.163} \\
\noalign{\hrule height 1pt}

\end{tabular}
\end{adjustbox}
\vspace{-0.5mm}
\caption{\textbf{Quantitative results of image dehazing on the HSTS subset of the RESIDE dataset.} The best results are boldfaced.}
\label{tab:dehazing}
\vspace{-5.0mm}
\end{table}

%% file: tab/denoising.tex
\begin{table}[tb]\small
\centering
\begin{adjustbox}{max width=\linewidth}
\begin{tabular}{l|ccc|ccc}
\noalign{\hrule height 1pt}
\multirow{2}{*}{ Methods} & \multicolumn{3}{c|}{Property{\rule{0pt}{1.1em}}} & \multicolumn{3}{c}{Kodak24} \\
\cline{2-7}
 & B{\rule{0pt}{1.1em}} & U & Z & PSNR$\uparrow$ & SSIM$\uparrow$ & LPIPS$\downarrow$ \\
\noalign{\hrule height 1pt}
\multicolumn{7}{c}{Task-specific} \\
\hline
C2N~\cite{Jang_2021_ICCV} & \ding{55} & \ding{55} & \ding{55} & \textbf{27.60} & \textbf{0.817} & \textbf{0.261} \\
MPI~\cite{ma2024masked} & \ding{55} & \ding{55} & \ding{51} & 27.33 & 0.768 & \textbf{0.261} \\
PUCA~\cite{jang2023puca} & \ding{55} & \ding{55} & \ding{55} & 27.12 & 0.776 & 0.282 \\
AP-BSN~\cite{Lee_2022_CVPR} & \ding{55} & \ding{55} & \ding{55} & 27.54 & 0.782 & 0.287 \\
\hline
\multicolumn{7}{c}{Posterior sampling} \\
\hline
GDP~\cite{Fei_2023_CVPR} & \ding{55} & \ding{51} & \ding{51} & 22.37 & 0.715 & 0.244 \\
TAO~\cite{gou2024test} & \ding{51} & \ding{51} & \ding{51} & 27.12 & 0.768 & 0.222 \\
LD-RPS~\cite{li2025ld} & \ding{51} & \ding{51} & \ding{51} & 27.66 & 0.830 & 0.176 \\
Ours & \ding{51} & \ding{51} & \ding{51} & \textbf{28.51} & \textbf{0.845} & \textbf{0.155} \\
\noalign{\hrule height 1pt}
\end{tabular}
\end{adjustbox}
\vspace{-0.5mm}
\caption{\textbf{Quantitative results of image denoising on the Kodak24 dataset.} The best results are highlighted in bold.}
\label{tab:denoising}
\vspace{-5.0mm}
\end{table}

%% file: tab/ablation.tex
\begin{table*}[htbp]\small
\centering
\begin{tabular}{l|ccccc|ccc|ccc}
\noalign{\hrule height 1pt}
\multirow{2}{*}{Methods} & \multicolumn{5}{c|}{LOLv1{\rule{0pt}{1.1em}}} & \multicolumn{3}{c|}{HSTS} & \multicolumn{3}{c}{Kodak24}\\
\cline{2-12}
& PSNR$\uparrow${\rule{0pt}{1.1em}} & SSIM$\uparrow$ & LPIPS$\downarrow$ & PI$\downarrow$ & NIQE$\downarrow$ & PSNR$\uparrow$ & SSIM$\uparrow$ & LPIPS$\downarrow$ & PSNR$\uparrow$ & SSIM$\uparrow$ & LPIPS$\downarrow$ \\
\noalign{\hrule height 1pt}
w/o PCDM & 17.78 & 0.807 & 0.290 & 5.03 & 5.48 & 21.05 & 0.809 & 0.177 & 28.00 & 0.840 & 0.166 \\
w/o DegLoss & 17.93 & 0.809 & 0.272 & 5.01 & 5.65 & 21.16 & 0.815 & 0.172 & 28.24 & 0.835 & 0.160 \\
w/o DQR & 17.92 & 0.811 & 0.271 & 4.88 & 5.50 & 21.22 & 0.813 & 0.175 & 28.16 & 0.844 & 0.155 \\
Full Version & \textbf{18.21} & \textbf{0.823} & \textbf{0.241} & \textbf{4.65} & \textbf{5.47} & \textbf{21.51} & \textbf{0.820} & \textbf{0.163} & \textbf{28.51} & \textbf{0.845} & \textbf{0.155} \\
\noalign{\hrule height 1pt}
\end{tabular}
\vspace{-2mm}
\caption{\textbf{Ablation results for key components across three restoration tasks.} The best outcomes are highlighted in bold.}
\label{tab:ablation}
\vspace{-4mm}
\end{table*}

%% file: tab/mixed.tex
\begin{table}[tbp]\small
\centering
\begin{adjustbox}{max width=\linewidth}
\begin{tabular}{l|ccc|ccc}
\noalign{\hrule height 1pt}
\multirow{2}{*}{Methods} & \multicolumn{3}{c|}{Low-light + Noise} & \multicolumn{3}{c}{Low-light + Haze + Noise}\\
\cline{2-7}
& PSNR$\uparrow$ & SSIM$\uparrow$ & LPIPS$\downarrow$ & PSNR$\uparrow$ & SSIM$\uparrow$ & LPIPS$\downarrow$ \\
\noalign{\hrule height 1pt}
GDP~\cite{Fei_2023_CVPR} & 16.27 & 0.669 & 0.272 & 16.31 & 0.660 & 0.278 \\
TAO~\cite{gou2024test} & 17.38 & 0.776 & 0.341 & 17.03 & 0.773 & 0.345 \\
LD-RPS~\cite{li2025ld} & 16.87 & 0.781 & 0.316 & 16.77 & 0.781 & 0.311 \\
Ours & \textbf{18.00} & \textbf{0.812} & \textbf{0.271} & \textbf{17.86} & \textbf{0.810} & \textbf{0.275} \\
\noalign{\hrule height 1pt}
\end{tabular}
\end{adjustbox}
\vspace{-1.5mm}
\caption{\textbf{Quantitative comparisons on two mixed degradation datasets.} The best results are indicated in bold.}
\label{tab:mixed}
\vspace{-3mm}
\end{table}

%% file: sec/6_Conclusion.tex
\section{Conclusion}

In this paper, a physically grounded zero-shot restoration framework named UP-ZeroIR is proposed, which unifies heterogeneous degradations as a compact distribution. With the physically coherent degradation modeling, distribution-alignment loss, and dynamic quality-refinement strategy, our method turns diffusion into a controllable search and produces physically consistent solutions across single and mixed degradations. Experiments show marked gains in accuracy and perception with reduced uncertainty from fixed-trajectory diffusion, and ablations confirm complementary benefits in representation strength, principled guidance, and convergence robustness. Notably, this is the first time to explore a physically interpretable image restoration paradigm via explicit distribution-level controllability, which is practical for zero-shot deployment in the real world. As a flexible pipeline, our diffusion model can adjust the parametric setting to robustly handle unseen real-world variability.

\noindent\textbf{Acknowledgements.} This work was supported by NSFC under Grants 62231002 and 62501027, and the Fundamental Research Funds for the Central Universities.

%% file: sec/X_suppl.tex
\clearpage
\setcounter{page}{1}
\maketitlesupplementary


\section{Additional Finding Analysis}

\subsection{Degradation Dataset Construction}
We constructed degraded datasets by introducing noise, haze, and low light to clean images from three datasets: Kodak24~\cite{franzen1999kodak}, the HSTS subset of RESIDE~\cite{li2018benchmarking}, and LOLv1~\cite{Wei2018RetinexNet}. The degradation processes are implemented as follows:
\begin{itemize}
    \item \textbf{Noise:} Gaussian noise is added to clean images which are normalized to the range \([0,1]\), as defined below:
    \begin{equation}
        x_{lq} = \text{clip}(x_{hq} + N(0, \sigma^2)),
    \end{equation}
    where \(\sigma\) controls the noise intensity (\emph{Light}: \(\sigma = 20/255\), \emph{Heavy}: \(\sigma = 50/255\)), and \(\text{clip}(\cdot)\) ensures that pixel values remain within the valid range [0, 1].
    
    \item \textbf{Haze:} hazy images are generated based on the atmospheric scattering model:
    \begin{equation}
        \mathbf{x}_{lq}(\mathbf{p}) = \mathbf{x}_{hq}(\mathbf{p}) \cdot t(\mathbf{p}) + \mathbf{A} \cdot (1 - t(\mathbf{p})),
    \end{equation}
    where \(t(\mathbf{p}) = \exp(-\beta \cdot d(\mathbf{p}))\). Here, \(\mathbf{A}\) denotes the global atmospheric light, which is fixed as \(\mathbf{A} = 1\) for all images, and \(\beta\) controls the haze density (\emph{Light}: \(\beta \in [0.02, 0.05]\), \emph{Heavy}: \(\beta \in [0.06, 0.09]\)). The variable \(d(\mathbf{p})\) represents a pseudo depth map that simulates the scene distance at pixel position \(\mathbf{p} = (i, j)\), defined as:
    \begin{equation}
        d(i,j) = \frac{ \sqrt{(i - c_x)^2 + (j - c_y)^2}} {s},
    \end{equation}
    where \((c_x, c_y)\) denotes the image center and \(s\) is the normalization scale. For images with height \(\text{row}\) and width \(\text{col}\), we set the scale as \(s = \sqrt{\max(\text{row}, \text{col})}\) and the center as \((c_x, c_y) = (\text{row}/2, \text{col}/2)\).

    \item \textbf{Low Light:} low-light images are generated by linearly scaling the brightness of clean images in the HSV color space. Given a clean image \(\mathbf{x}_{hq}\) with HSV channels \((\mathbf{H}, \mathbf{S}, \mathbf{V})\), the low-light image \(\mathbf{x}_{lq}\) is obtained as:
    \begin{equation}
        \begin{aligned}
            \mathbf{V}_{lq} &= \text{clip}(\gamma \cdot \mathbf{V}_{hq}),\\
            \mathbf{x}_{lq} &= \text{HSV2RGB}(\mathbf{H}, \mathbf{S}, \mathbf{V}_{lq}),
        \end{aligned}
    \end{equation}
    where \(\gamma \in (0,1)\) is a scaling factor controlling the illumination level. Specifically, \(\gamma = 0.7\) for light degradation and \(\gamma = 0.3\) for heavy degradation.
\end{itemize}

\begin{figure}[htbp]
    \centering
    \begin{subfigure}{1.0\linewidth}
        \centering
        \includegraphics[trim={1mm 1mm 1mm 1mm}, clip, width=\linewidth]{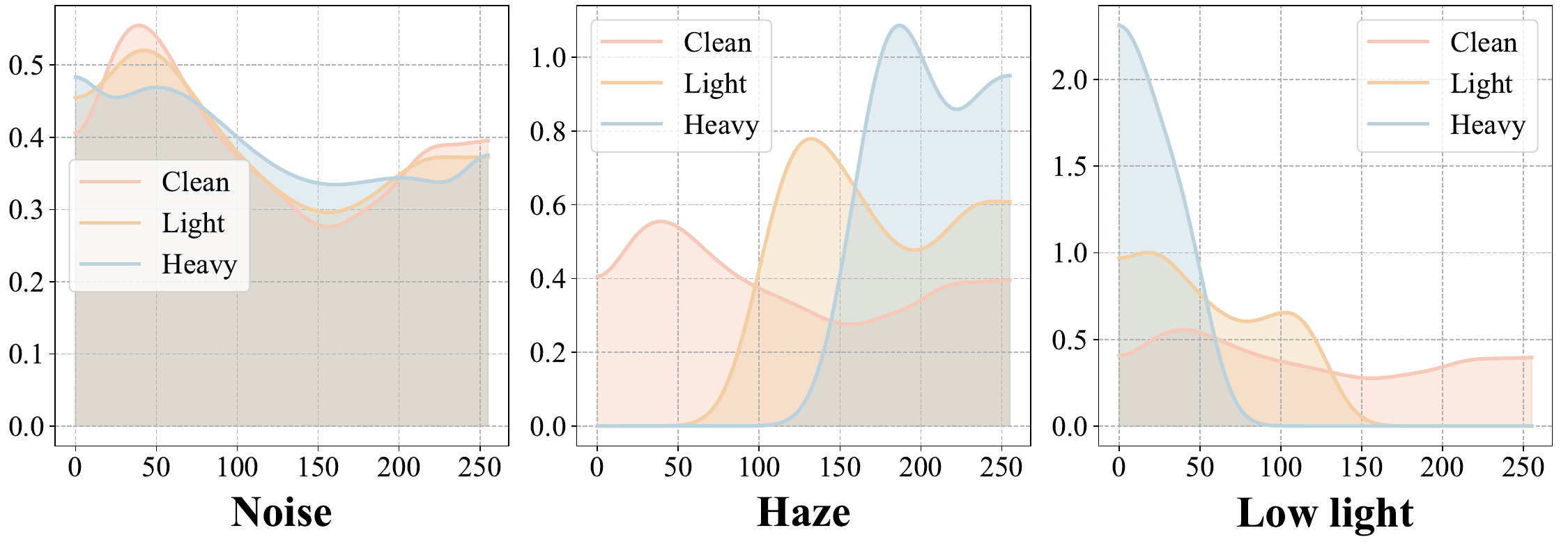}
        \caption{Distribution of the HSTS dataset for image dehazing.}
    \end{subfigure}
    \begin{subfigure}{1.0\linewidth}
        \centering
        \includegraphics[trim={1mm 1mm 1mm 1mm}, clip, width=\linewidth]{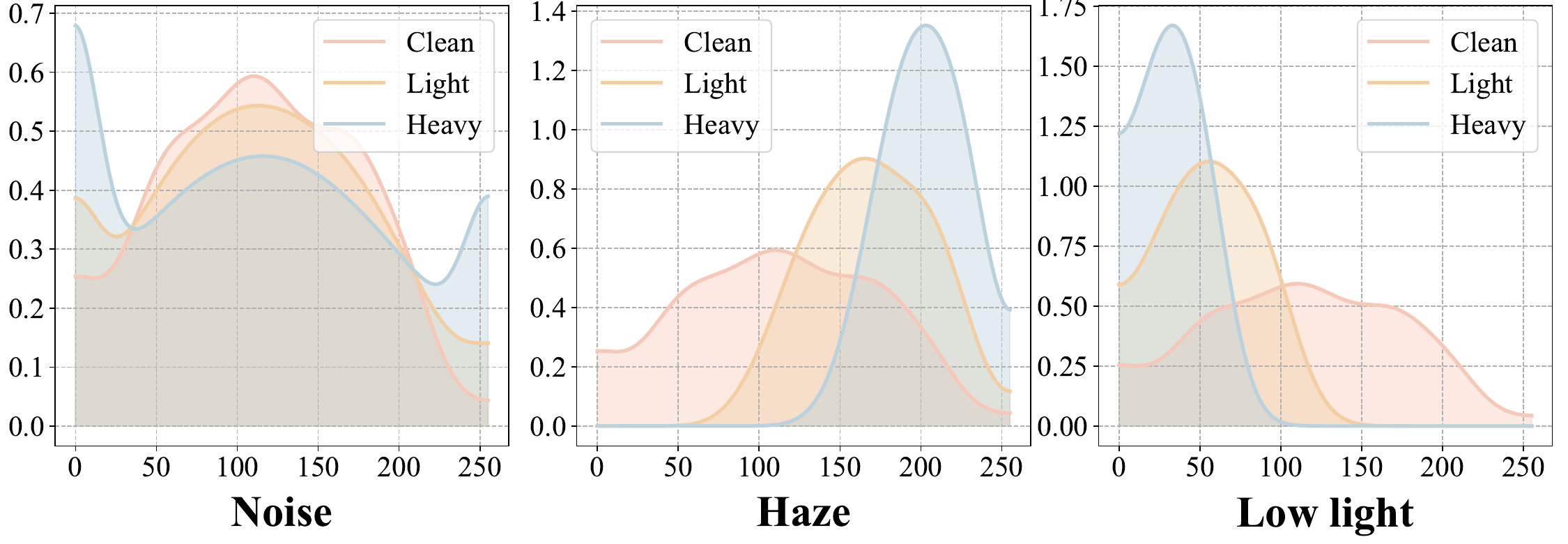}
        \caption{Distribution of the LOLv1 dataset for low-light enhancement.}
    \end{subfigure}
    \caption{\textbf{Pixel-domain distributions across various degradation types and intensities.} The x-axis and y-axis represent pixel values and probability density, respectively. Each degradation type is analyzed across three intensities: clean, light, and heavy.}
    \label{fig:baseline_degradation}
\end{figure}
\begin{figure*}[htbp]
    \centering
    \begin{subfigure}{1.0\linewidth}
        \centering
        \includegraphics[trim={1mm 5mm 1mm 1mm}, clip, width=\linewidth]{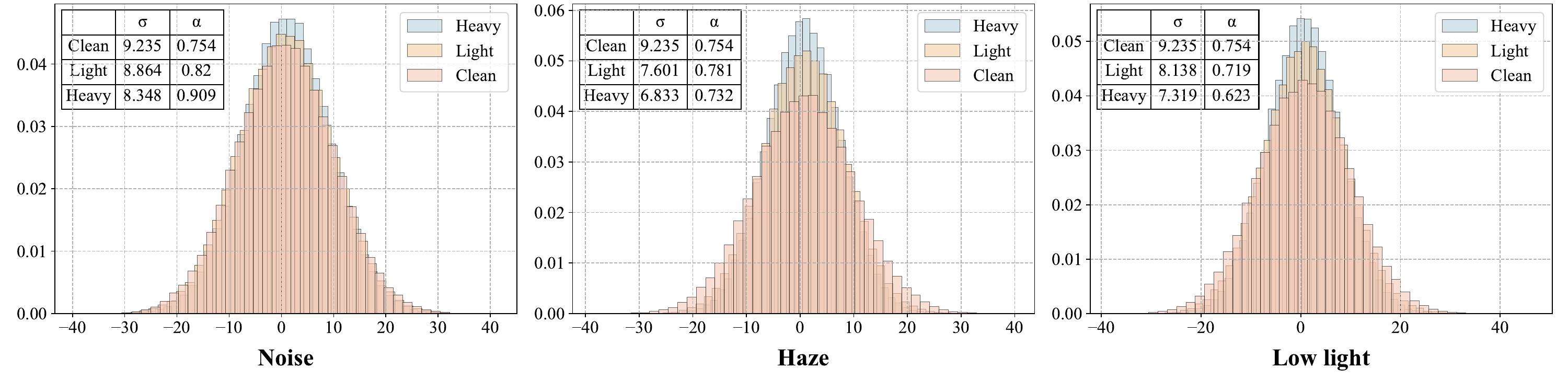}
        \caption{Distribution of the HSTS dataset for image dehazing.}
    \end{subfigure}
    \begin{subfigure}{1.0\linewidth}
        \centering
        \includegraphics[trim={1mm 5mm 1mm 1mm}, clip, width=\linewidth]{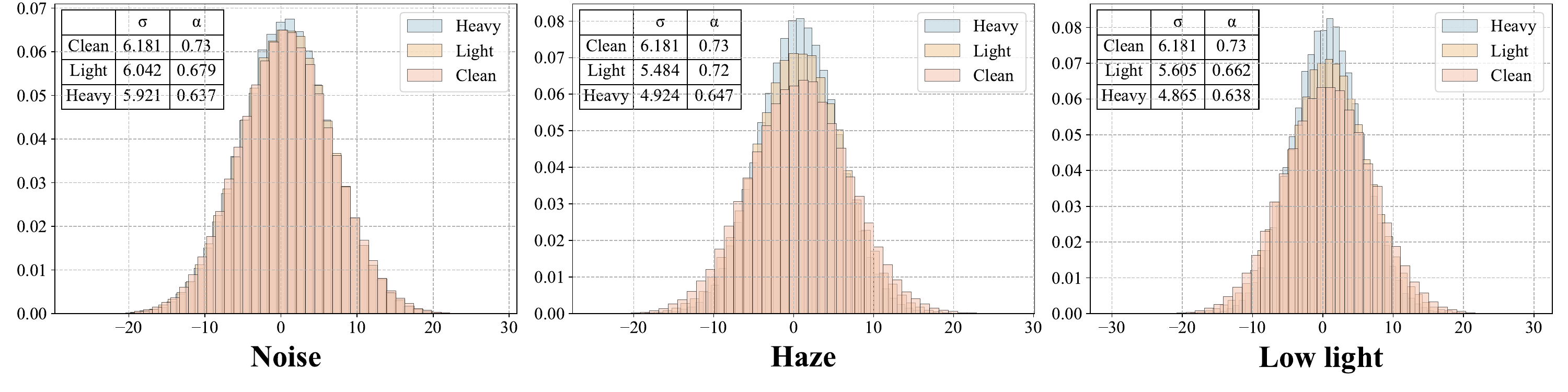}
        \caption{Distribution of the LOLv1 dataset for low-light enhancement.}
    \end{subfigure}
    \caption{\textbf{Homogeneous degradation distribution.}
    The x-axis represents the latent representation, while the y-axis denotes the probability density. Parameters \(\sigma\) and \(\alpha\) define the dispersion and tail behavior of the distribution, respectively.}
    \label{fig:latent_degradation}
\end{figure*}
\subsection{Visualization Results on Other Datasets}
In the main paper, we focused on statistical analyses on the Kodak24 dataset.
This section extends our observations to the HSTS subset of RESIDE and the LOLv1 datasets.
As shown in \cref{fig:baseline_degradation}, heterogeneous degradations induce consistently strong divergences and systematic shifts in pixel-domain distributions across all datasets due to the distinct underlying mechanisms.
\cref{fig:latent_degradation} demonstrates that our physically coherent degradation representation provides a unified and physically grounded parameterization of degraded images. This enables homogeneous modeling of heterogeneous degradations and further supports the consistency of the findings reported in the main paper.

\begin{table}[t]
    \begin{adjustbox}{max width=\linewidth}
    \begin{tabular}{c|cccccc}
    \noalign{\hrule height 1pt}
    Task{\rule{0pt}{1.1em}} & \(\lambda_1\) & \(\lambda_2\) & \(\lambda_3\) & \(\lambda_4\) & \(\lambda_5\) & \(\lambda_6\) \\
    \noalign{\hrule height 1pt}
    Low-light enhancement & 1e-5 & 1e0 & 2e-3 & 5e-5 & 1e-2 & 1e-2 \\
    Image dehazing & 1e-3 & 1e0 & 5e-3 & 5e-5 & 1e-2 & 1e-3 \\
    Image denoising & 1e-3 & 1e0 & 5e-4 & 5e-5 & 1e-5 & 1e-5 \\
    \noalign{\hrule height 1pt}
    \end{tabular}
    \end{adjustbox}
    \caption{\textbf{Loss weighting factors for different restoration tasks.}}
    \label{tab:loss_weighting}
\end{table}

\begin{table*}[t]
    \centering
    \begin{adjustbox}{max width=\textwidth}
    \begin{tabular}{l|ccc|ccccc|ccccc}
    \noalign{\hrule height 1pt}
    \multirow{2}{*}{Methods} & \multicolumn{3}{c|}{Property{\rule{0pt}{1.1em}}} & \multicolumn{5}{c|}{LOLv1} & \multicolumn{5}{c}{LOLv2} \\
    \cline{2-14} 
    & B{\rule{0pt}{1.1em}} & U & Z & PSNR$\uparrow$ & SSIM$\uparrow$ & LPIPS$\downarrow$ & PI$\downarrow$ & NIQE$\downarrow$ & PSNR$\uparrow$ & SSIM$\uparrow$ & LPIPS$\downarrow$ & PI$\downarrow$ & NIQE$\downarrow$ \\
    \noalign{\hrule height 1pt}
    \multicolumn{14}{c}{Supervised All-in-one} \\
    \hline
    AirNet~\cite{Li_2022_CVPR} & \ding{51} & \ding{55} & \ding{55} & 7.07 &  0.121 & 0.813 & 8.06 & 8.29 & 9.62 & 0.181 & 0.534 & 8.13 & 9.30 \\
    PromptIR~\cite{potlapalli2023promptir} & \ding{51} & \ding{55} & \ding{55} & 7.73 &  0.171 & 0.549 & 9.70 & 11.09 & 9.73 & 0.189 & 0.534 & 9.61 & 11.61 \\
    DiffUIR~\cite{Zheng_2024_CVPR} & \ding{51} & \ding{55} & \ding{55} & \textbf{21.36} & \textbf{0.907} & \textbf{0.125} & \textbf{4.68} & \textbf{5.95} & \textbf{26.14} & \textbf{0.898} & \textbf{0.114} & \textbf{5.26} & \textbf{7.34} \\
    \hline
    \multicolumn{14}{c}{Posterior sampling} \\
    \hline
    GDP~\cite{Fei_2023_CVPR} & \ding{55} & \ding{51} & \ding{51} & 16.86 & 0.689 & 0.299 & 4.67 & 5.91 & 15.32 & 0.597 & 0.337 & 5.13 & 8.18 \\
    TAO~\cite{gou2024test} & \ding{51} & \ding{51} & \ding{51} & 17.42 & 0.792 & 0.320 & 6.03 & 7.74 & 16.78 & 0.747 & \textbf{0.314} & 6.41 & 9.44 \\
    LD-RPS~\cite{li2025ld} & \ding{51} & \ding{51} & \ding{51} & 17.26 & 0.797 & 0.291 & 5.02 & 5.79 & 18.22 & 0.744 & 0.335 & 5.05 & 6.03 \\
    Ours & \ding{51} & \ding{51} & \ding{51} & \textbf{18.21} & \textbf{0.823} & \textbf{0.241} & \textbf{4.65} & \textbf{5.47} & \textbf{19.20} & \textbf{0.761} & 0.332 & \textbf{4.98} & \textbf{5.72} \\
    \noalign{\hrule height 1pt}
    \end{tabular}
    \end{adjustbox}
    \caption{\textbf{Quantitative results of low-light enhancement on the LOLv1 and LOLv2 datasets.} The best results within each category are boldfaced. Methods are grouped by three properties: B (task-blind), U (unsupervised), and Z (zero-shot), reflecting their adaptability.}
    \label{tab:lowlightE_supp}
\end{table*}
\begin{figure}[t]
    \centering
    \includegraphics[trim={15mm 13mm 11mm 11mm}, clip, width=\linewidth]{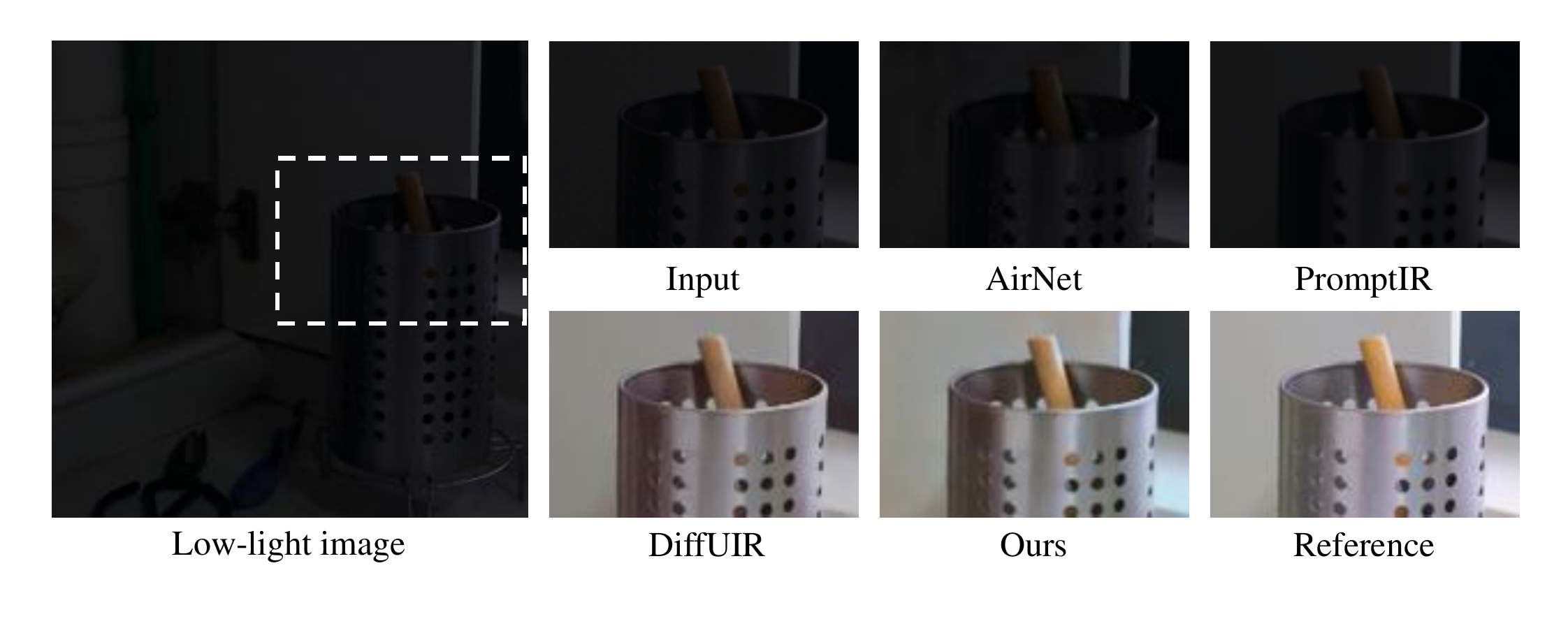}
    \caption{\textbf{Qualitative performance of low-light enhancement on the LOLv1 dataset, corresponding to the main paper.}}
    \label{fig:result_lowlight_main}
    \vspace{-1mm}
\end{figure}
\section{More Implementation Details}
In all experiments, we adopt the pre-trained Stable Diffusion model and set the timestep to \(T = 1000\).
The diffusion inference process is divided into three stages.
In the first stage, \( t \in \left [ 1000, 700 \right ) \), we train PCDM exclusively using the Adam optimizer with a learning rate of \(1\times 10^{-5}\), enabling accurately modeling the physically coherent degradation transformation.
During the second stage, \( t \in \left [ 700, 150 \right ) \), we update the posterior sampling trajectory by incorporating the degradation alignment loss \(J_{deg}\) along with additional image distance terms.
In the final stage, \( t \in \left [ 150, 0 \right ) \), we further introduce the image quality term \(Q\) to enhance the perceptual fidelity of the restored output.
The corresponding loss weighting factors, which are fine-tuned for each restoration task to ensure optimal parameter settings, are summarized in \cref{tab:loss_weighting}.

For the dynamic quality-refinement strategy, the refinement depth is set to an intermediate diffusion step, specifically $t' = 500$.
In the mixed-degradation experiments, we construct datasets by introducing light-level atmospheric-scattering haze and Gaussian noise into the low-light images of the LOLv1 dataset, following the degradation procedures described above.
For a comprehensive and fair evaluation, we include three categories of comparison baselines: supervised all-in-one methods, task-specific methods, and zero-shot posterior-sampling methods.

\begin{table}[t]
    \centering
    \begin{adjustbox}{max width=\linewidth}
    \begin{tabular}{l|ccc|ccc}
    \noalign{\hrule height 1pt}
    \multirow{2}{*}{Methods} & \multicolumn{3}{c|}{Property{\rule{0pt}{1.1em}}} & \multicolumn{3}{c}{HSTS} \\
    \cline{2-7}
     & B{\rule{0pt}{1.1em}} & U & Z & PSNR$\uparrow$ & SSIM$\uparrow$ & LPIPS$\downarrow$ \\
    \noalign{\hrule height 1pt}
    \multicolumn{7}{c}{Supervised All-in-one} \\
    \hline
    AirNet~\cite{Li_2022_CVPR}& \ding{51} & \ding{55} & \ding{55} & 24.37 & 0.899 & 0.059 \\
    PromptIR~\cite{potlapalli2023promptir} & \ding{51} & \ding{55} & \ding{55} & 25.67 & 0.907 & 0.048 \\
    DiffUIR~\cite{Zheng_2024_CVPR} & \ding{51} & \ding{55} & \ding{55} & \textbf{26.88} & \textbf{0.914}  & \textbf{0.045} \\
    \hline
    \multicolumn{7}{c}{Posterior sampling} \\
    \hline
    GDP~\cite{Fei_2023_CVPR} & \ding{55} & \ding{51} & \ding{51} & 12.57 & 0.703 & 0.164 \\
    TAO~\cite{gou2024test} & \ding{51} & \ding{51} & \ding{51} & 15.66 & 0.775 & 0.216 \\
    LD-RPS~\cite{li2025ld} & \ding{51} & \ding{51} & \ding{51} & 20.48 & 0.804 & 0.173 \\
    Ours & \ding{51} & \ding{51} & \ding{51} & \textbf{21.51} & \textbf{0.820} & \textbf{0.163} \\
    \noalign{\hrule height 1pt}
    \end{tabular}
    \end{adjustbox}
    \caption{\textbf{Quantitative results of image dehazing on the HSTS subset of the RESIDE dataset.} The best results are boldfaced.}
    \label{tab:dehazing_supp}
\end{table}
\begin{figure}[t]
    \centering
    \includegraphics[trim={8mm 6mm 6mm 16mm}, clip, width=\linewidth]{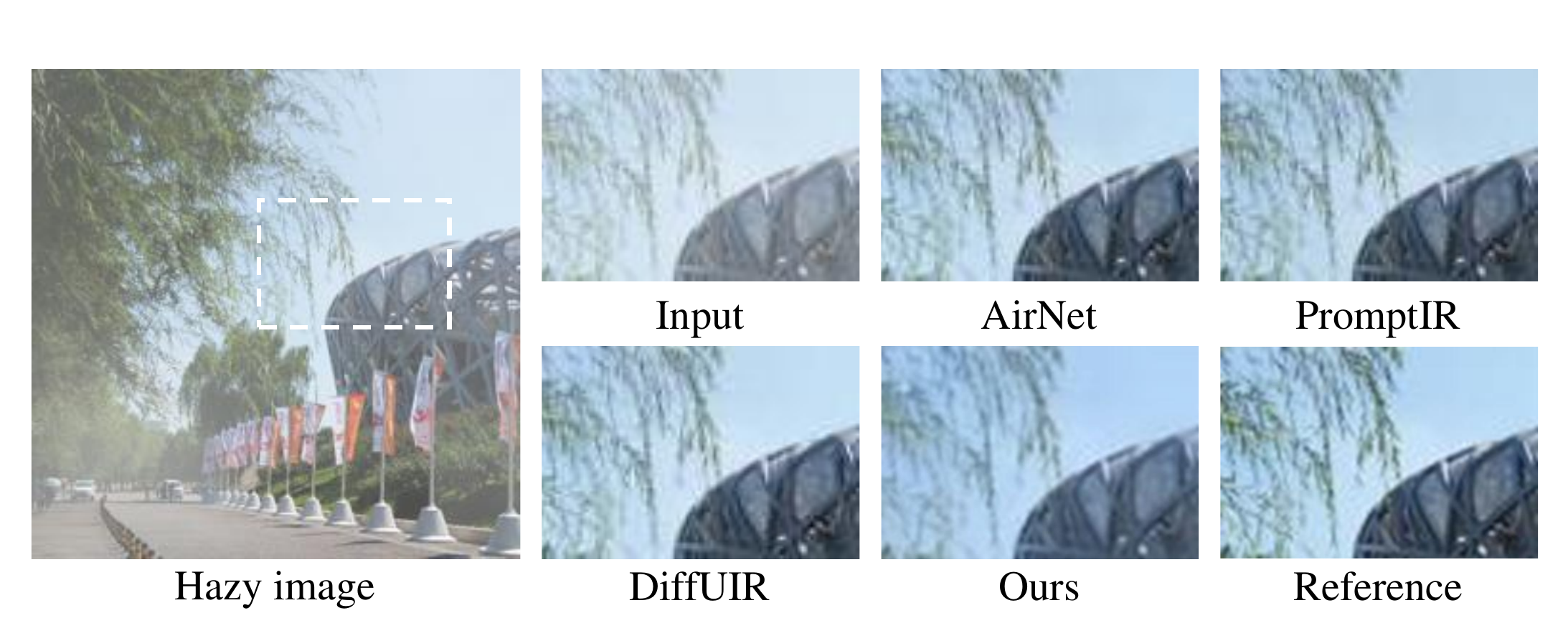}
    \caption{\textbf{Qualitative performance of image dehazing on the HSTS dataset, corresponding to the main paper.}}
    \label{fig:result_dehazing_main}
\end{figure}
\section{Additional Experimental Results}
\begin{figure}[t]
    \centering
    \includegraphics[trim={8mm 6mm 6mm 16mm}, clip, width=\linewidth]{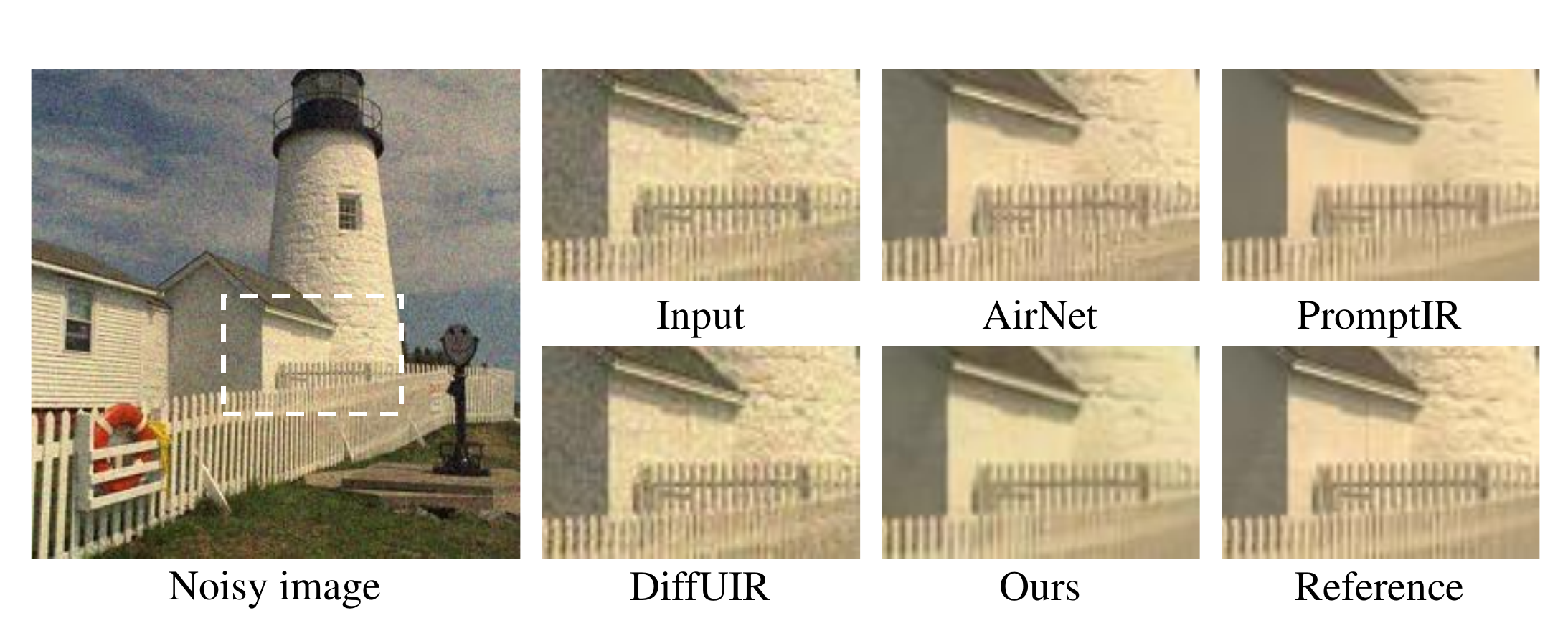}
    \caption{\textbf{Qualitative performance of image denoising on the Kodak24 dataset, corresponding to the main paper.}}
    \label{fig:result_denoising_main}
\end{figure}

\begin{figure*}[t]
    \centering
    \begin{subfigure}{\linewidth}
        \centering
        \includegraphics[trim={28mm 5mm 28mm 5mm}, clip, width=\linewidth]{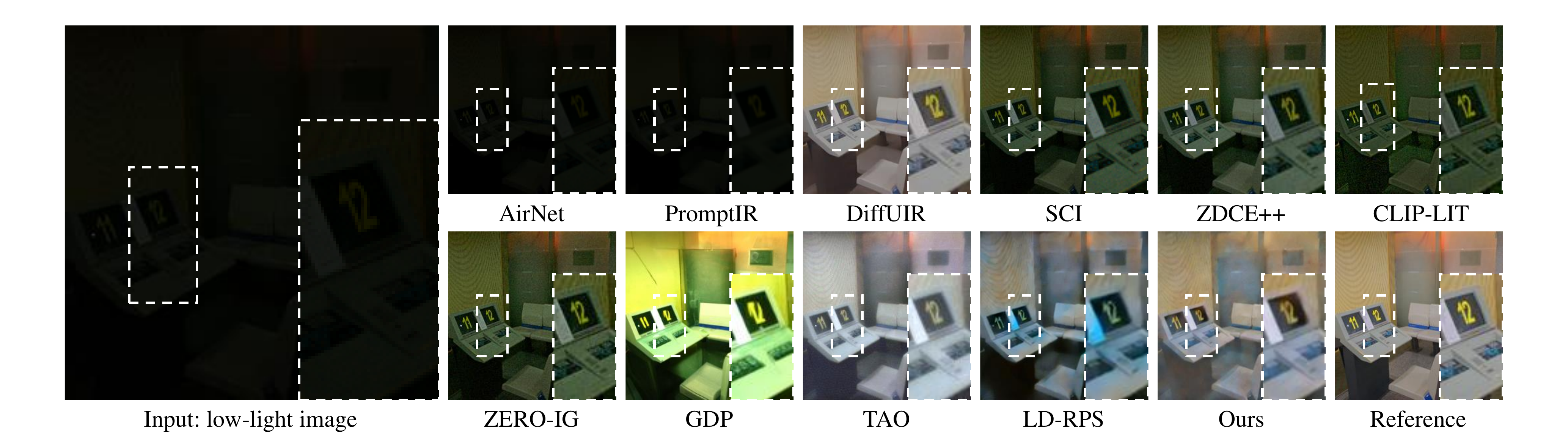}
        \caption{LOLv1: 665}
    \end{subfigure}
    \begin{subfigure}{\linewidth}
        \centering
        \includegraphics[trim={28mm 5mm 28mm 5mm}, clip, width=\linewidth]{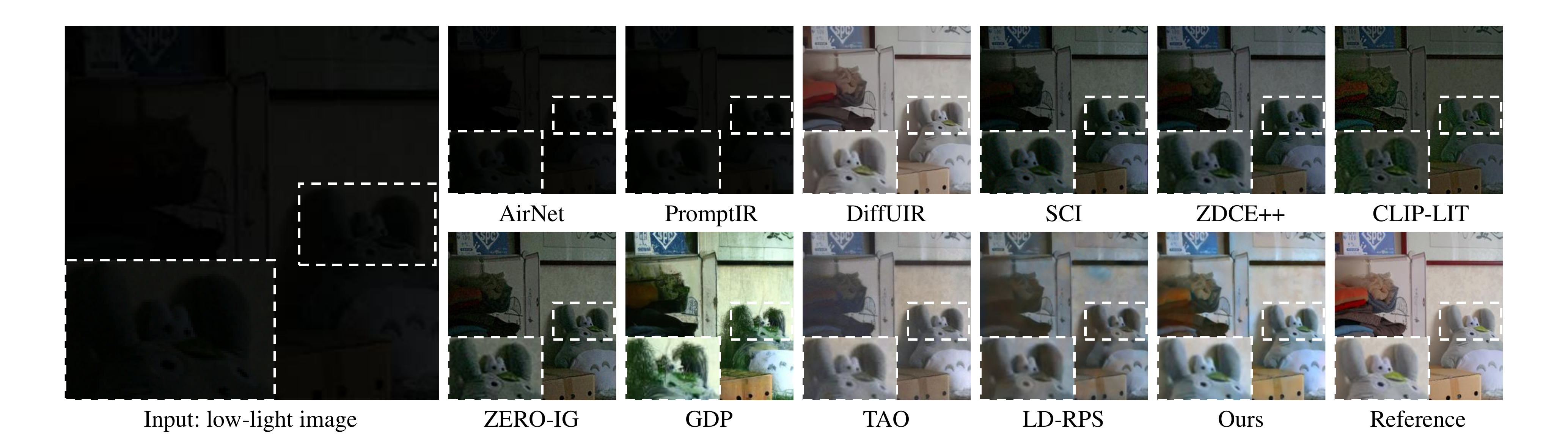}
        \caption{LOLv1: 23}
    \end{subfigure}
    \begin{subfigure}{\linewidth}
        \centering
        \includegraphics[trim={28mm 5mm 28mm 5mm}, clip, width=\linewidth]{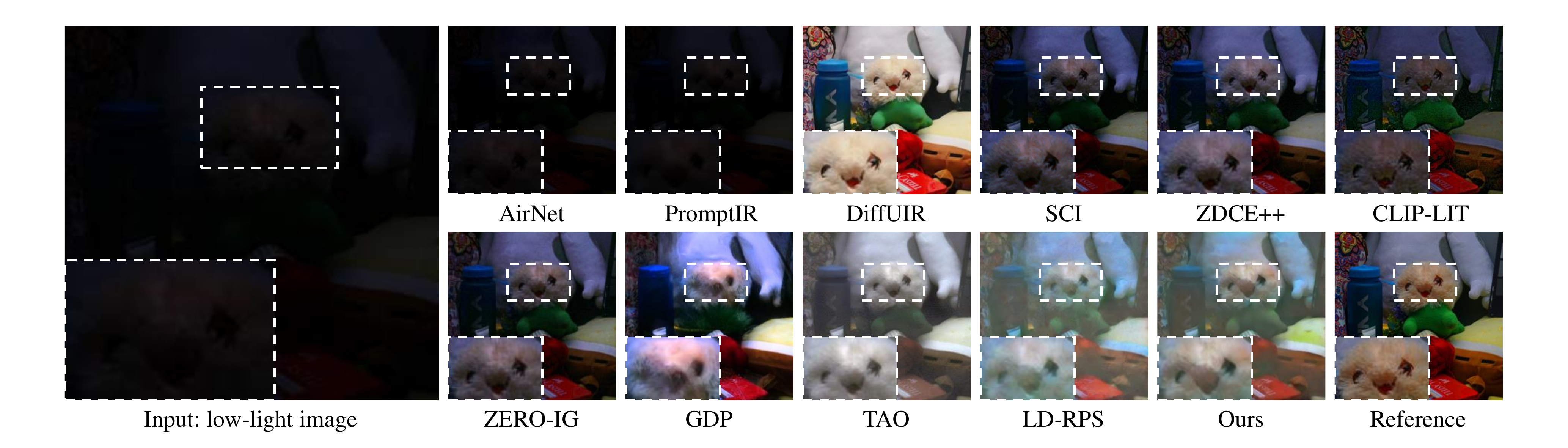}
        \caption{LOLv1: 493}
    \end{subfigure}
    \caption{\textbf{Supplementary qualitative results of low-light enhancement on the LOLv1 dataset.}}
    \label{fig:result_lowlight_supp}
\end{figure*}

\begin{figure*}[t]
    \centering
    \begin{subfigure}{\linewidth}
        \centering
        \includegraphics[trim={28mm 5mm 28mm 5mm}, clip, width=\linewidth]{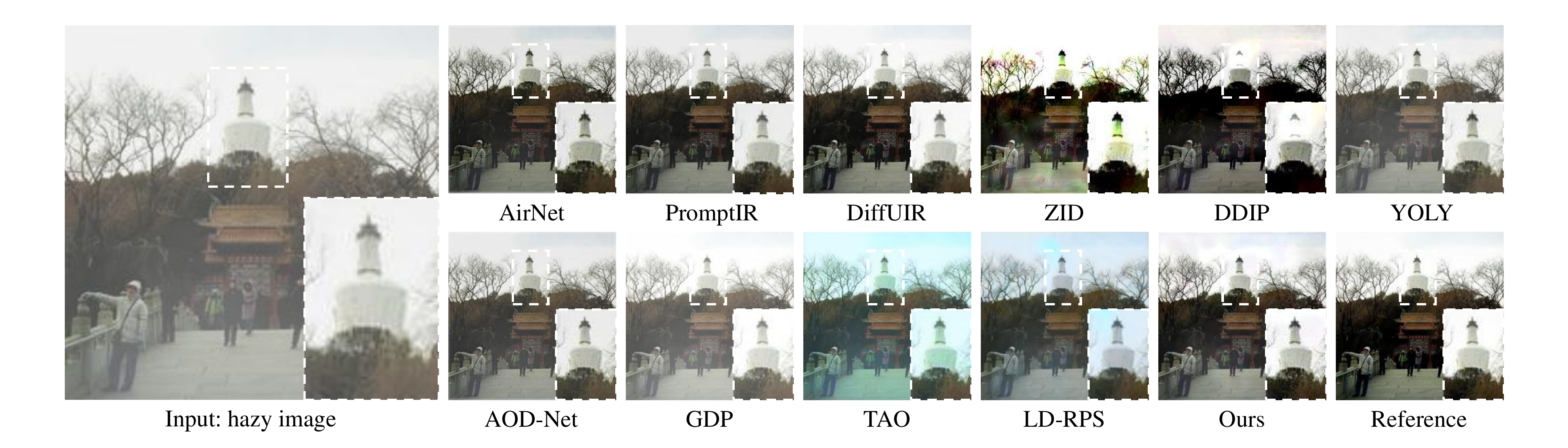}
        \caption{HSTS: 5576}
    \end{subfigure}
    \begin{subfigure}{\linewidth}
        \centering
        \includegraphics[trim={28mm 5mm 28mm 5mm}, clip, width=\linewidth]{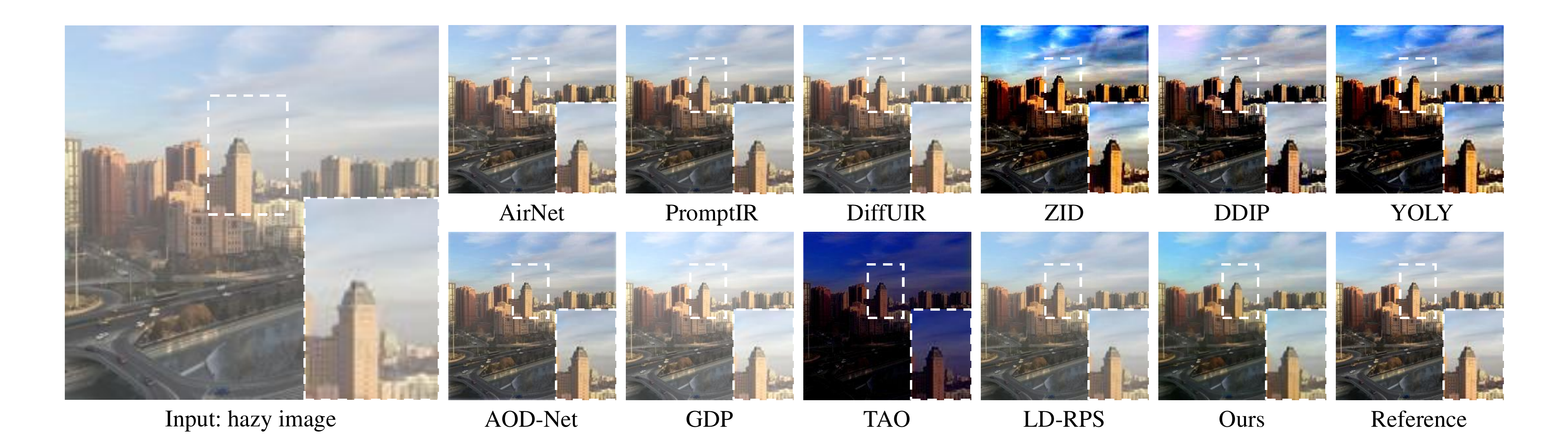}
        \caption{HSTS: 5920}
    \end{subfigure}
    \begin{subfigure}{\linewidth}
        \centering
        \includegraphics[trim={28mm 5mm 28mm 5mm}, clip, width=\linewidth]{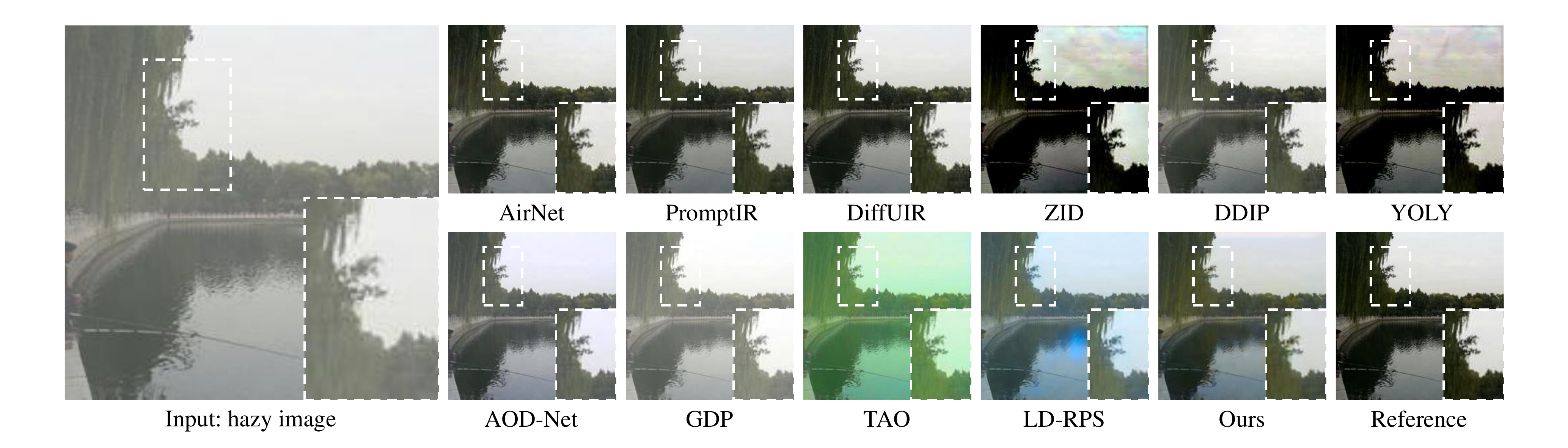}
        \caption{HSTS: 3146}
    \end{subfigure}
    \vspace{-7mm}
    \caption{\textbf{Supplementary qualitative performance of image dehazing on the HSTS dataset.}}
    \label{fig:result_dehazing_supp}
    \vspace{-5mm}
\end{figure*}

\begin{figure*}[t]
    \centering
    \begin{subfigure}{\linewidth}
        \centering
        \includegraphics[trim={28mm 5mm 28mm 5mm}, clip, width=\linewidth]{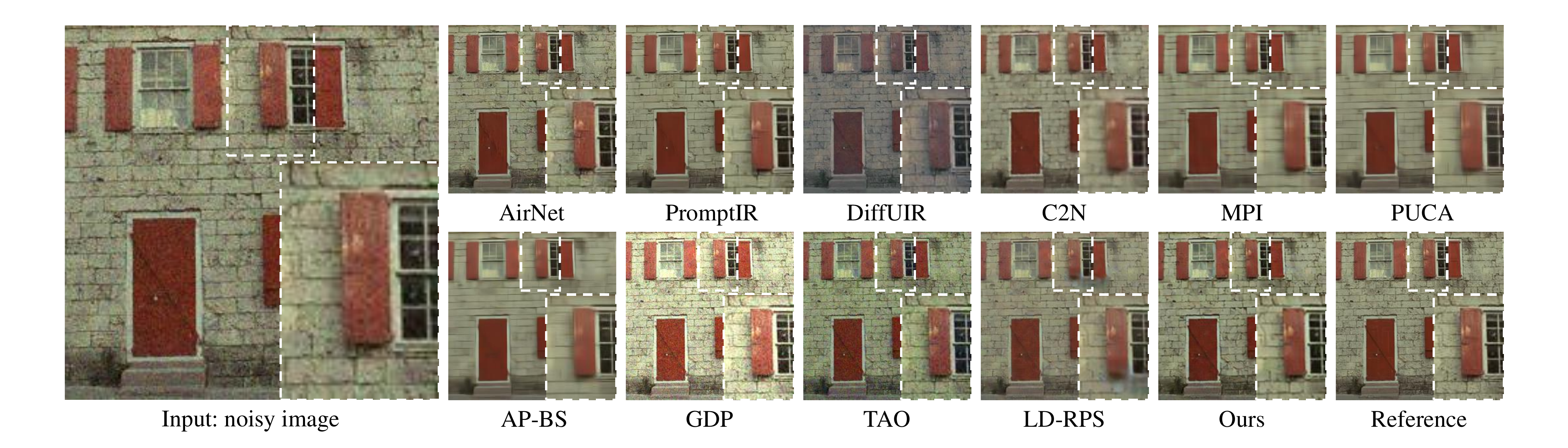}
        \caption{Kodak24: kodim01}
    \end{subfigure}
    \begin{subfigure}{\linewidth}
        \centering
        \includegraphics[trim={28mm 5mm 28mm 5mm}, clip, width=\linewidth]{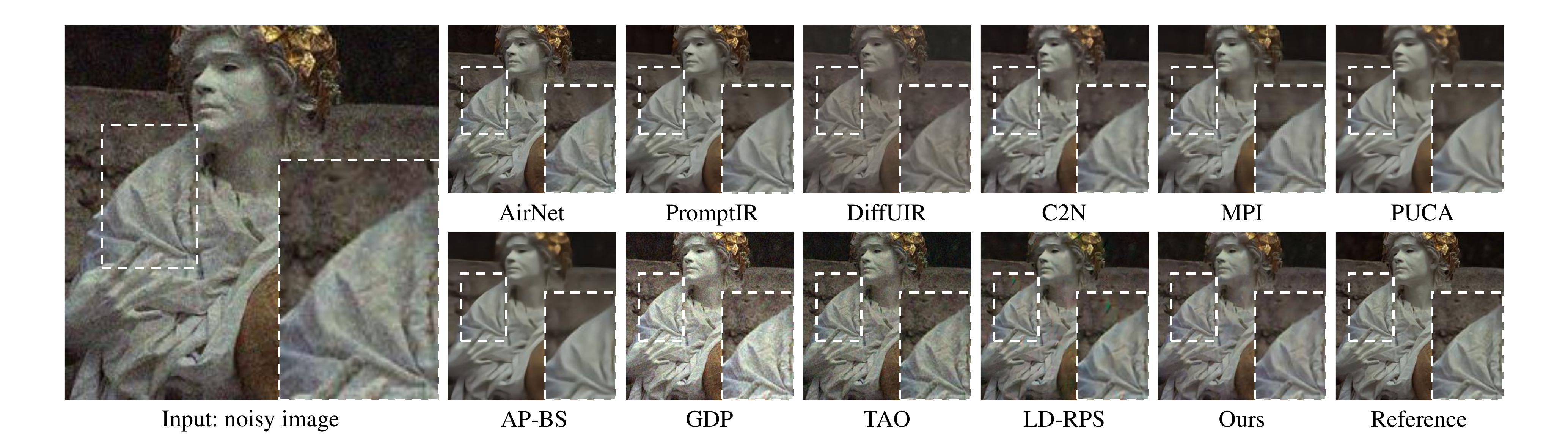}
        \caption{Kodak24: kodim17}
    \end{subfigure}
    \begin{subfigure}{\linewidth}
        \centering
        \includegraphics[trim={28mm 5mm 28mm 5mm}, clip, width=\linewidth]{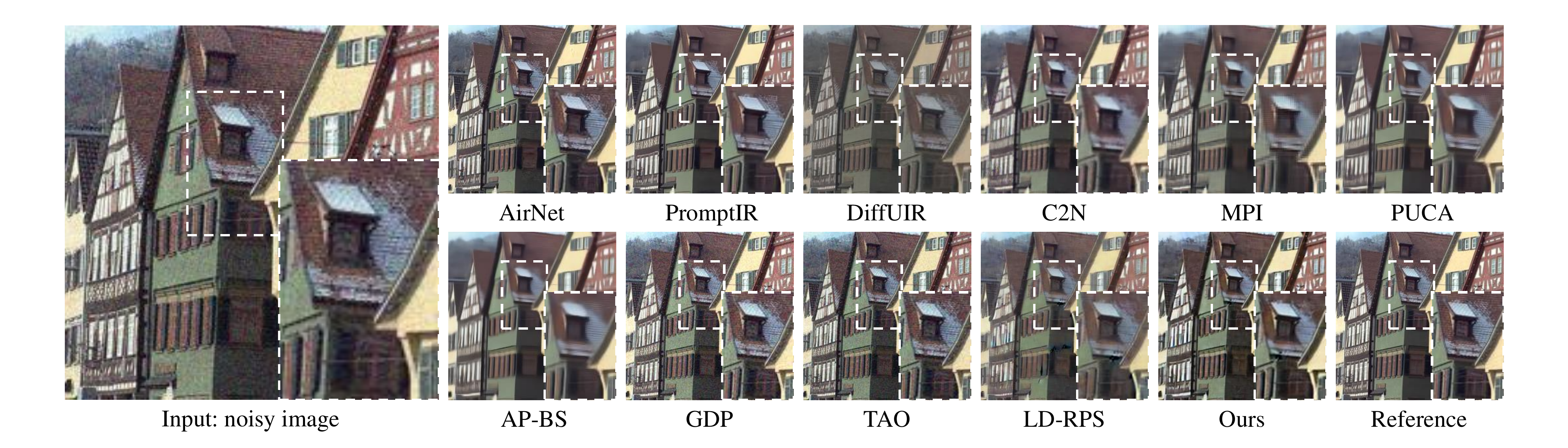}
        \caption{Kodak24: kodim08}
    \end{subfigure}
    \caption{\textbf{Supplementary qualitative performance of image denoising on the Kodak24 dataset.}}
    \label{fig:result_denoising_supp}
\end{figure*}

\noindent\textbf{Single Degradation.}
The main paper presents comparisons of our method with task-specific and posterior sampling approaches on low-light enhancement, image dehazing, and image denoising tasks. In this section, we provide the evaluation results for supervised all-in-one methods, including AirNet~\cite{Li_2022_CVPR}, PromptIR~\cite{potlapalli2023promptir}, and DiffUIR~\cite{Zheng_2024_CVPR}, which correspond to the results reported in the main paper. As shown in \crefrange{tab:lowlightE_supp}{tab:denoising_supp}, our method achieves quantitative results comparable to these supervised all-in-one methods on several metrics.
Qualitative results in \crefrange{fig:result_lowlight_main}{fig:result_denoising_main} illustrate that our method delivers visual fidelity comparable to, and even surpassing, that of supervised all-in-one models.
We further provide additional visual comparisons in \crefrange{fig:result_lowlight_supp}{fig:result_denoising_supp}.
Compared with other baselines, our method achieves state-of-the-art performance while avoid common adverse effects, such as over-smoothing, color distortions, and structural loss, thereby effectively balancing between degradation removal and visual fidelity.

Moreover, due to the limited generalization ability of supervised methods, they often struggle to handle unseen degradations during training.
The poor performance of AirNet and PromptIR on low-light enhancement, as well as the suboptimal results of DiffUIR on image denoising, further highlight the superiority of our zero-shot framework.

\begin{figure*}[t]
    \centering
    \begin{subfigure}{0.49\linewidth}
        \centering
        \includegraphics[trim={2mm 2mm 2mm 2mm}, clip, width=\linewidth]{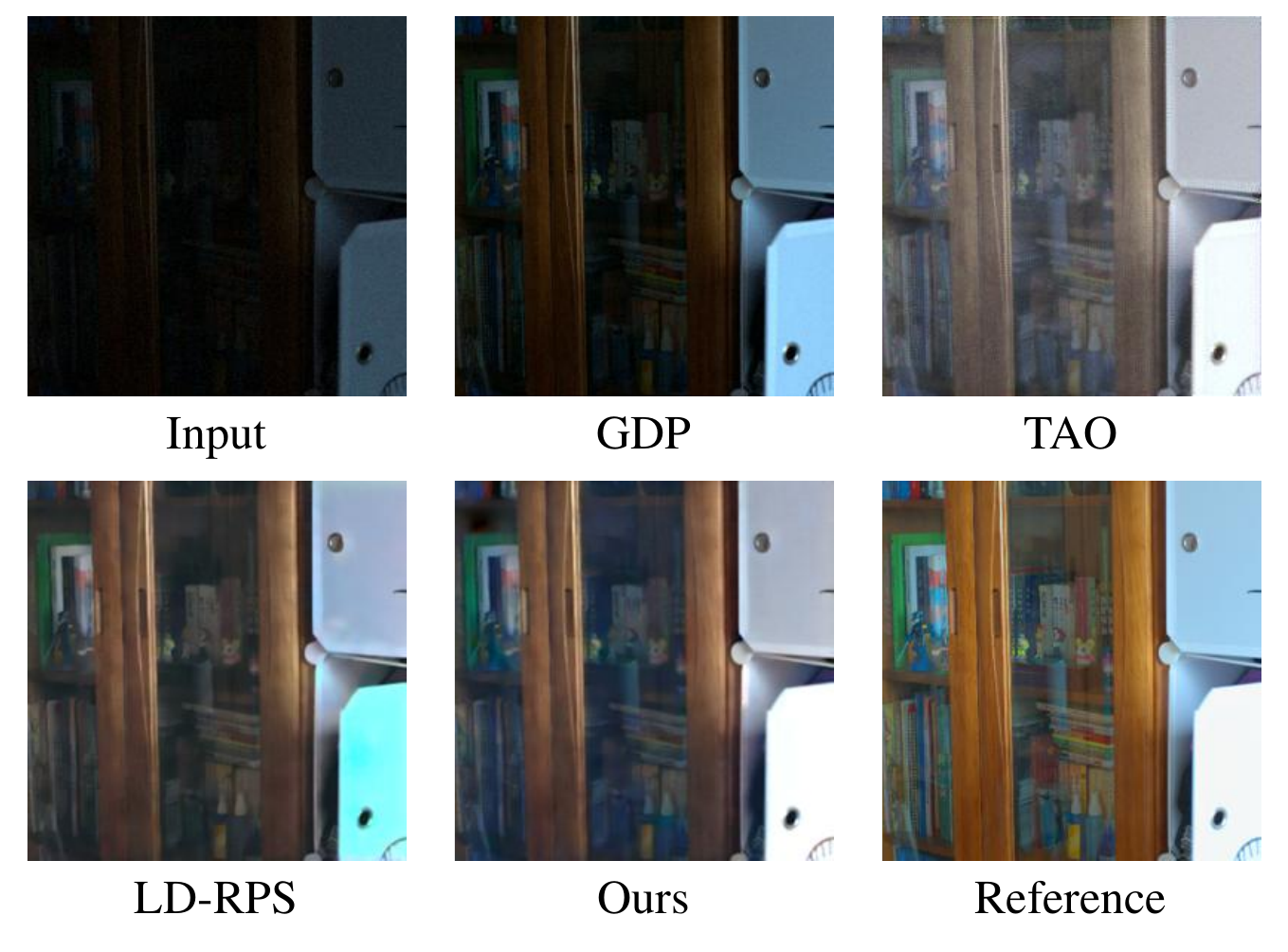}
        \caption{LOLv1: 1}
    \end{subfigure}
    \begin{subfigure}{0.49\linewidth}
        \centering
        \includegraphics[trim={2mm 2mm 2mm 2mm}, clip, width=\linewidth]{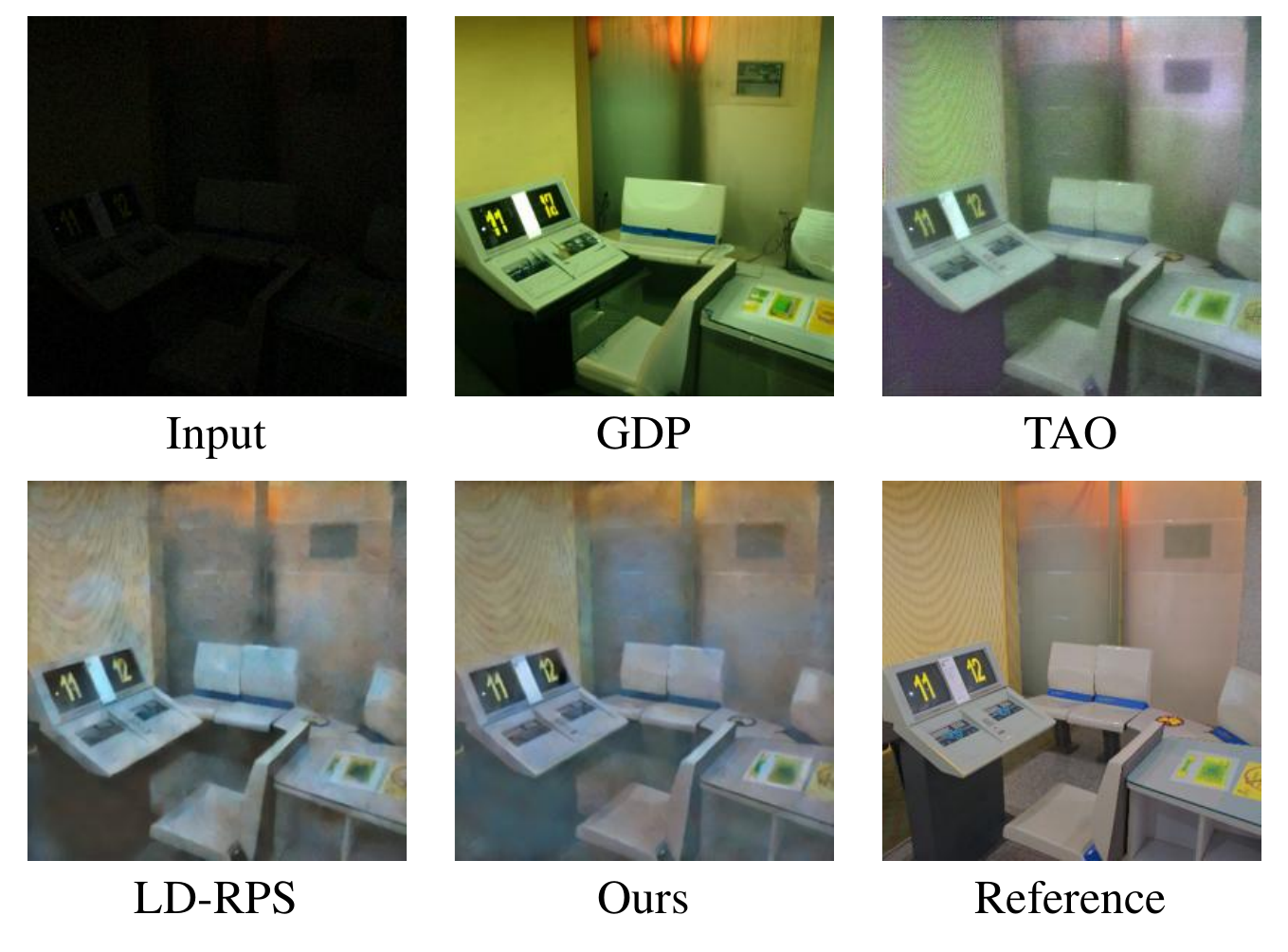}
        \caption{LOLv1: 665}
    \end{subfigure}
    \caption{\textbf{Visual comparisons on the Low-light + Noise scenario.}}
    \label{fig:mixed_degradation2}
\end{figure*}

\begin{figure*}[t]
    \centering
    \begin{subfigure}{0.49\linewidth}
        \centering
        \includegraphics[trim={2mm 2mm 2mm 2mm}, clip, width=\linewidth]{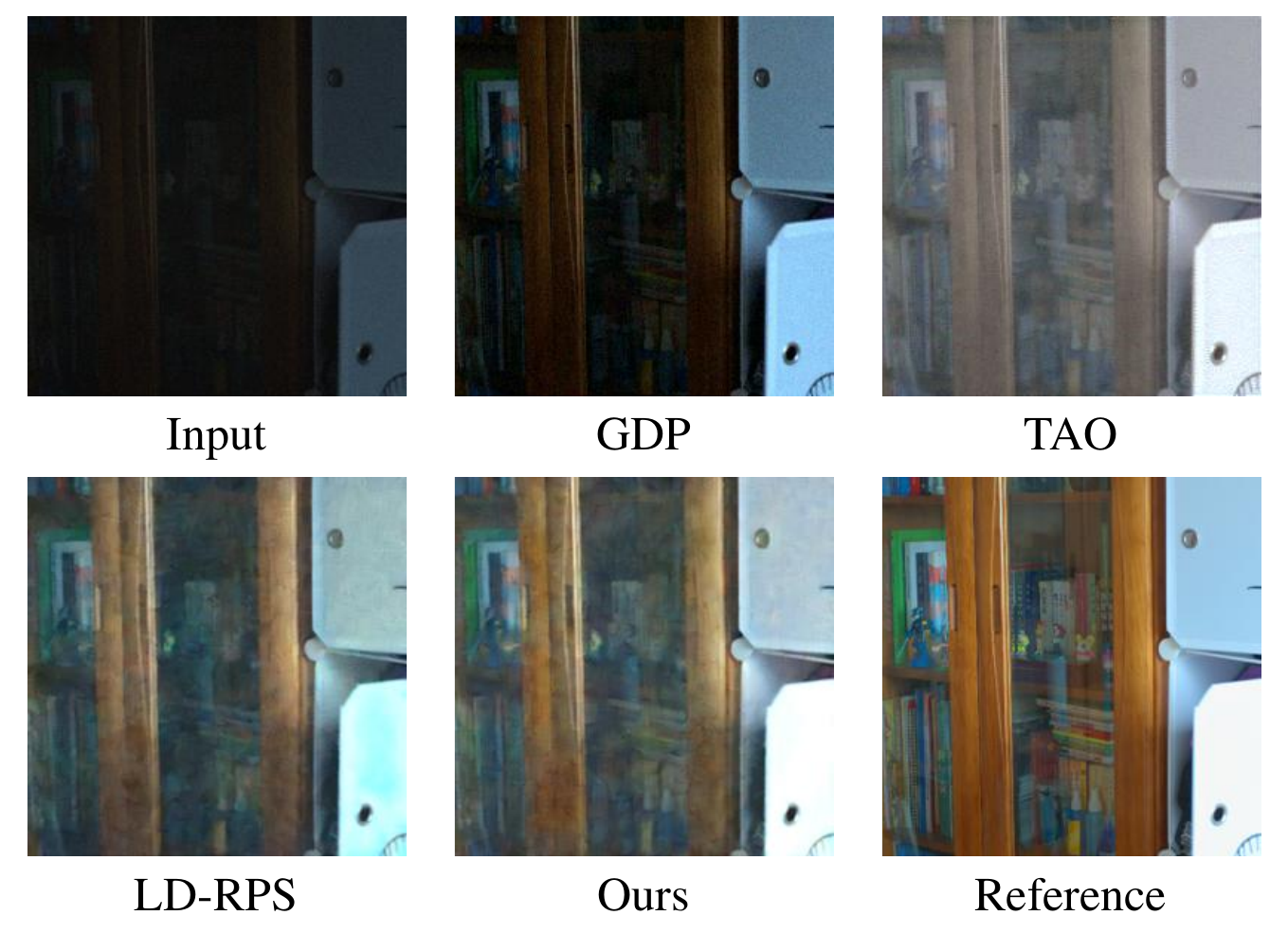}
        \caption{LOLv1: 1}
    \end{subfigure}
    \begin{subfigure}{0.49\linewidth}
        \centering
        \includegraphics[trim={2mm 2mm 2mm 2mm}, clip, width=\linewidth]{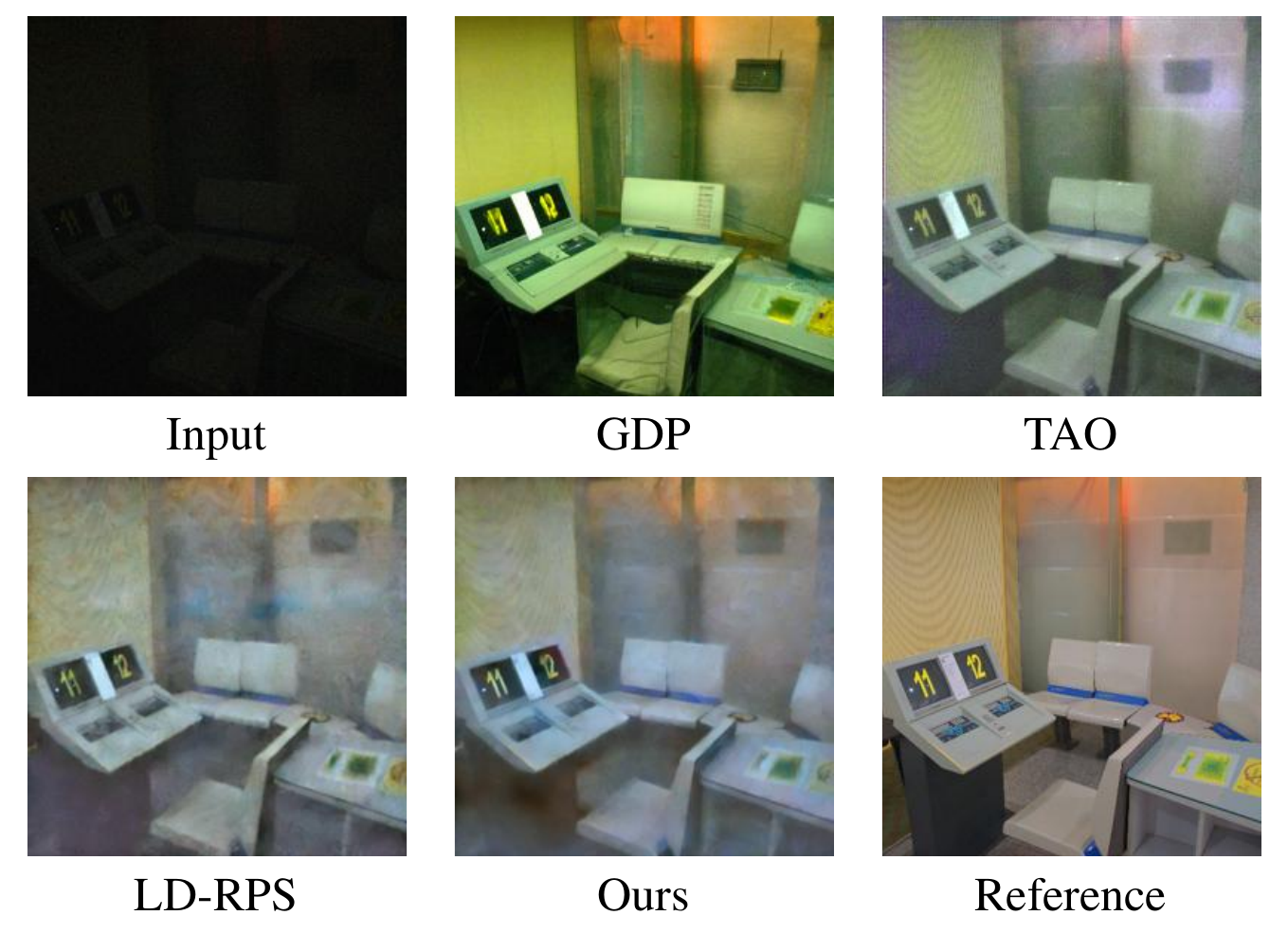}
        \caption{LOLv1: 665}
    \end{subfigure}
    \caption{\textbf{Visual comparisons on the Low-light + Haze + Noise scenario.}}
    \label{fig:mixed_degradation3}
\end{figure*}

\noindent\textbf{Mixed Degradation.}
To intuitively illustrate the robustness under mixed degradations, we present the visual results of our UP-ZeroIR alongside those of other posterior-sampling methods.
As shown in \cref{fig:mixed_degradation2} and \cref{fig:mixed_degradation3}, our method simultaneously handles low light, noise, and haze under the combined effects of multiple degradations, significantly improving image quality and naturalness for robust restoration.

\begin{table}[t]
    \centering
    \begin{adjustbox}{max width=\linewidth}
    \begin{tabular}{l|ccc|ccc}
    \noalign{\hrule height 1pt}
    \multirow{2}{*}{ Methods} & \multicolumn{3}{c|}{Property{\rule{0pt}{1.1em}}} & \multicolumn{3}{c}{Kodak24} \\
    \cline{2-7}
     & B{\rule{0pt}{1.1em}} & U & Z & PSNR$\uparrow$ & SSIM$\uparrow$ & LPIPS$\downarrow$ \\
    \noalign{\hrule height 1pt}
    \multicolumn{7}{c}{Supervised All-in-one} \\
    \hline
    AirNet~\cite{Li_2022_CVPR} & \ding{51} & \ding{55} & \ding{55} & 29.94 & 0.834 & 0.114 \\
    PromptIR~\cite{potlapalli2023promptir} & \ding{51} & \ding{55} & \ding{55} & \textbf{30.88} & \textbf{0.873} & \textbf{0.113} \\
    DiffUIR~\cite{Zheng_2024_CVPR} & \ding{51} & \ding{55} & \ding{55} & 22.86 & 0.789 & 0.219 \\
    \hline
    \multicolumn{7}{c}{Posterior sampling} \\
    \hline
    GDP~\cite{Fei_2023_CVPR} & \ding{55} & \ding{51} & \ding{51} & 22.37 & 0.715 & 0.244 \\
    TAO~\cite{gou2024test} & \ding{51} & \ding{51} & \ding{51} & 27.12 & 0.768 & 0.222 \\
    LD-RPS~\cite{li2025ld} & \ding{51} & \ding{51} & \ding{51} & 27.66 & 0.830 & 0.176 \\
    Ours & \ding{51} & \ding{51} & \ding{51} & \textbf{28.51} & \textbf{0.845} & \textbf{0.155} \\
    \noalign{\hrule height 1pt}
    \end{tabular}
    \end{adjustbox}
    \caption{\textbf{Quantitative results of image denoising on the Kodak24 dataset.} The best results are highlighted in bold.}
    \label{tab:denoising_supp}
\end{table}
\section{Discussion}
\noindent\textbf{Model Efficiency.}
Compared with supervised methods, our method eliminates the computational burden of training and reduces the reliance on paired datasets, thereby substantially lowering the overall resource consumption.
Our method achieves a lightweight design with only 1M parameters and leverages a physically coherent degradation model to adaptively handle various degradations.
For a fair comparison, we evaluate our approach against the state-of-the-art LD-RPS, as both methods are built on the pre-trained Stable Diffusion model.
With the proposed dynamic quality-refinement strategy, our method facilitates a more efficient diffusion inference process, achieving an approximate 30\% reduction in inference time.

\noindent\textbf{Task Adaptability.}
Our method demonstrates strong generalizability and can be effectively transferred to other image restoration tasks.
As analyzed in the main paper, heterogeneous degradations can be universally parameterized as a physically consistent distribution in the latent diffusion space.
During the diffusion inference, our method adaptively optimizes the degradation distribution of the input image, progressively converging to the clean image, regardless of the type of degradation.
This enables our UP-ZeroIR to adapt to a variety of restoration tasks while maintaining superior and robust restoration performance.
Furthermore, our proposed physically coherent degradation model operates as a plug-and-play solution, demonstrating its significant potential for unified degradation modeling and alignment across a variety of image-related tasks.




%% file: main.bib
@String(CVPR= {IEEE Conf. Comput. Vis. Pattern Recog.})

@String(ICCV= {Int. Conf. Comput. Vis.})

@String(ECCV= {Eur. Conf. Comput. Vis.})

@String(BMVC= {Brit. Mach. Vis. Conf.})

@String(ICLR = {Int. Conf. Learn. Represent.})

@String(CVPR  = {CVPR})

@String(ICCV  = {ICCV})

@String(ECCV  = {ECCV})

@String(BMVC  =	{BMVC})

@String(ICLR  = {ICLR})

@article{IRsurvey,
  author={Jiang, Junjun and Zuo, Zengyuan and Wu, Gang and Jiang, Kui and Liu, Xianming},
  journal={IEEE Transactions on Pattern Analysis and Machine Intelligence}, 
  title={A Survey on All-in-One Image Restoration: Taxonomy, Evaluation and Future Trends}, 
  year={2025},
  volume={47},
  number={12},
  pages={11892-11911},
  doi={10.1109/TPAMI.2025.3598132}}

@ARTICLE{Digitalsurvey,
  author={Andrews, Harry C.},
  journal={Computer}, 
  title={Digital image restoration: A survey}, 
  year={1974},
  volume={7},
  number={5},
  pages={36-45},
  doi={10.1109/MC.1974.6323527},
  ISSN={1558-0814},
  month={May},}

@article{seshadrinathan2010study,
  title={Study of subjective and objective quality assessment of video},
  author={Seshadrinathan, Kalpana and Soundararajan, Rajiv and Bovik, Alan Conrad and Cormack, Lawrence K},
  journal={IEEE transactions on Image Processing},
  volume={19},
  number={6},
  pages={1427--1441},
  year={2010},
  publisher={IEEE}
}

@article{Dabov2007BM3D,
  title={Image denoising by sparse 3-D transform-domain collaborative filtering},
  author={Dabov, Kostadin and Foi, Alessandro and Katkovnik, Vladimir and Egiazarian, Karen},
  journal={IEEE Trans. Image Processing},
  volume={16},
  number={8},
  pages={2080--2095},
  year={2007}
}

@inproceedings{Zhang2017DnCNN,
  title={Beyond a Gaussian denoiser: Residual learning of deep CNN for image denoising},
  author={Zhang, Kai and Zuo, Wangmeng and Chen, Yunjin and Meng, Deyu and Zhang, Lei},
  booktitle={CVPR},
  pages={3929--3937},
  year={2017}
}

@inproceedings{Lim2017EDSR,
  title={Enhanced deep residual networks for single image super-resolution},
  author={Lim, Bee and Son, Sanghyun and Kim, Heewon and Nah, Seungjun and Lee, Kyoung Mu},
  booktitle={CVPR Workshops},
  pages={136--144},
  year={2017}
}

@article{Foi2008PG,
  title   = {Practical Poissonian-Gaussian noise modeling and fitting for single-image raw-data},
  author  = {Foi, Alessandro and Trimeche, Mohamed and Katkovnik, Vladimir and Egiazarian, Karen},
  journal = {IEEE Transactions on Image Processing},
  volume  = {17},
  number  = {10},
  pages   = {1737--1754},
  year    = {2008}
}

@inproceedings{Fattal2008SIGGRAPH,
  title     = {Single image dehazing},
  author    = {Fattal, Raanan},
  booktitle = {ACM SIGGRAPH},
  year      = {2008}
}

@inproceedings{Wei2018RetinexNet,
  title     = {Deep Retinex Decomposition for Low-Light Enhancement},
  author    = {Wei, Chen and Wang, Wenjing and Yang, Wenhan and Liu, Jiaying},
  booktitle = {BMVC},
  year      = {2018}
}

@inproceedings{Levin2009BlindDeconv,
  title     = {Understanding and Evaluating Blind Deconvolution Algorithms},
  author    = {Levin, Anat and Weiss, Yair and Durand, Fr{\'e}do and Freeman, William T.},
  booktitle = {Proceedings of the IEEE Conference on Computer Vision and Pattern Recognition (CVPR)},
  year      = {2009},
  pages     = {1964--1971}
}

@article{tan2015video,
  title={Video quality evaluation methodology and verification testing of HEVC compression performance},
  author={Tan, Thiow Keng and Weerakkody, Rajitha and Mrak, Marta and Ramzan, Naeem and Baroncini, Vittorio and Ohm, Jens-Rainer and Sullivan, Gary J},
  journal={IEEE Transactions on Circuits and Systems for Video Technology},
  volume={26},
  number={1},
  pages={76--90},
  year={2015},
  publisher={IEEE}
}

@inproceedings{Fu_2021_ICCV,
    author    = {Fu, Xueyang and Wang, Xi and Liu, Aiping and Han, Junwei and Zha, Zheng-Jun},
    title     = {Learning Dual Priors for JPEG Compression Artifacts Removal},
    booktitle = {Proceedings of the IEEE/CVF International Conference on Computer Vision (ICCV)},
    month     = {October},
    year      = {2021},
    pages     = {4086-4095}
}

@inproceedings{Parmar_2022_CVPR,
    author    = {Parmar, Gaurav and Zhang, Richard and Zhu, Jun-Yan},
    title     = {On Aliased Resizing and Surprising Subtleties in GAN Evaluation},
    booktitle = {Proceedings of the IEEE/CVF Conference on Computer Vision and Pattern Recognition (CVPR)},
    month     = {June},
    year      = {2022},
    pages     = {11410-11420}
}

@article{li2018structure,
  title={Structure-revealing low-light image enhancement via robust retinex model},
  author={Li, Mading and Liu, Jiaying and Yang, Wenhan and Sun, Xiaoyan and Guo, Zongming},
  journal={IEEE transactions on image processing},
  volume={27},
  number={6},
  pages={2828--2841},
  year={2018},
  publisher={IEEE}
}

@inproceedings{Shyam_2023_ICCV,
    author    = {Shyam, Pranjay and Yoo, HyunJin},
    title     = {Data Efficient Single Image Dehazing via Adversarial Auto-Augmentation and Extended Atmospheric Scattering Model},
    booktitle = {Proceedings of the IEEE/CVF International Conference on Computer Vision (ICCV) Workshops},
    month     = {October},
    year      = {2023},
    pages     = {227-237}
}

@inproceedings{Chen_2023_CVPR,
    author    = {Chen, Xiang and Li, Hao and Li, Mingqiang and Pan, Jinshan},
    title     = {Learning a Sparse Transformer Network for Effective Image Deraining},
    booktitle = {Proceedings of the IEEE/CVF Conference on Computer Vision and Pattern Recognition (CVPR)},
    month     = {June},
    year      = {2023},
    pages     = {5896-5905}
}

@inproceedings{Cho_2021_ICCV,
    author    = {Cho, Sung-Jin and Ji, Seo-Won and Hong, Jun-Pyo and Jung, Seung-Won and Ko, Sung-Jea},
    title     = {Rethinking Coarse-To-Fine Approach in Single Image Deblurring},
    booktitle = {Proceedings of the IEEE/CVF International Conference on Computer Vision (ICCV)},
    month     = {October},
    year      = {2021},
    pages     = {4641-4650}
}

@article{he2010single,
  title={Single image haze removal using dark channel prior},
  author={He, Kaiming and Sun, Jian and Tang, Xiaoou},
  journal={IEEE transactions on pattern analysis and machine intelligence},
  volume={33},
  number={12},
  pages={2341--2353},
  year={2010},
  publisher={IEEE}
}

@inproceedings{Liu_2024_CVPR,
    author    = {Liu, Jiawei and Wang, Qiang and Fan, Huijie and Wang, Yinong and Tang, Yandong and Qu, Liangqiong},
    title     = {Residual Denoising Diffusion Models},
    booktitle = {Proceedings of the IEEE/CVF Conference on Computer Vision and Pattern Recognition (CVPR)},
    month     = {June},
    year      = {2024},
    pages     = {2773-2783}
}

@inproceedings{Zamir_2022_CVPR,
    author    = {Zamir, Syed Waqas and Arora, Aditya and Khan, Salman and Hayat, Munawar and Khan, Fahad Shahbaz and Yang, Ming-Hsuan},
    title     = {Restormer: Efficient Transformer for High-Resolution Image Restoration},
    booktitle = {Proceedings of the IEEE/CVF Conference on Computer Vision and Pattern Recognition (CVPR)},
    month     = {June},
    year      = {2022},
    pages     = {5728-5739}
}

@article{xing2023daqe,
  title={Daqe: Enhancing the quality of compressed images by exploiting the inherent characteristic of defocus},
  author={Xing, Qunliang and Xu, Mai and Deng, Xin and Guo, Yichen},
  journal={IEEE Transactions on Pattern Analysis and Machine Intelligence},
  volume={45},
  number={8},
  pages={9611--9626},
  year={2023},
  publisher={IEEE}
}

@article{tang2025degradation,
  title={Degradation-aware residual-conditioned optimal transport for unified image restoration},
  author={Tang, Xiaole and Gu, Xiang and He, Xiaoyi and Hu, Xin and Sun, Jian},
  journal={IEEE Transactions on Pattern Analysis and Machine Intelligence},
  year={2025},
  publisher={IEEE}
}

@inproceedings{cui2025adair,
  title={Adair: Adaptive all-in-one image restoration via frequency mining and modulation},
  author={Cui, Yuning and Zamir, Syed Waqas and Khan, Salman and Knoll, Alois and Shah, Mubarak and Khan, Fahad Shahbaz},
  booktitle={Proceedings of the International Conference on Learning Representations (ICLR)},
  pages={57335--57356},
  year={2025}
}

@article{chrysos2020rocgan,
  title={Rocgan: Robust conditional gan},
  author={Chrysos, Grigorios G and Kossaifi, Jean and Zafeiriou, Stefanos},
  journal={International Journal of Computer Vision},
  volume={128},
  number={10},
  pages={2665--2683},
  year={2020},
  publisher={Springer}
}

@inproceedings{Li_2018_CVPR,
author = {Li, Runde and Pan, Jinshan and Li, Zechao and Tang, Jinhui},
title = {Single Image Dehazing via Conditional Generative Adversarial Network},
booktitle = {Proceedings of the IEEE Conference on Computer Vision and Pattern Recognition (CVPR)},
month = {June},
year = {2018}
}

@article{pan2020physics,
  title={Physics-based generative adversarial models for image restoration and beyond},
  author={Pan, Jinshan and Dong, Jiangxin and Liu, Yang and Zhang, Jiawei and Ren, Jimmy and Tang, Jinhui and Tai, Yu-Wing and Yang, Ming-Hsuan},
  journal={IEEE transactions on pattern analysis and machine intelligence},
  volume={43},
  number={7},
  pages={2449--2462},
  year={2020},
  publisher={IEEE}
}

@article{jiang2025msfa,
  title={MSFA Image Denoising Using Physics-based Noise Model and Noise-decoupled Network},
  author={Jiang, Yuqi and Fu, Ying and Liu, Qiankun and Zhang, Jun},
  journal={IEEE Transactions on Pattern Analysis and Machine Intelligence},
  year={2025},
  publisher={IEEE}
}

@inproceedings{Zhang_2023_CVPR,
    author    = {Zhang, Jinghao and Huang, Jie and Yao, Mingde and Yang, Zizheng and Yu, Hu and Zhou, Man and Zhao, Feng},
    title     = {Ingredient-Oriented Multi-Degradation Learning for Image Restoration},
    booktitle = {Proceedings of the IEEE/CVF Conference on Computer Vision and Pattern Recognition (CVPR)},
    month     = {June},
    year      = {2023},
    pages     = {5825-5835}
}

@inproceedings{jiang2024autodir,
  title={Autodir: Automatic all-in-one image restoration with latent diffusion},
  author={Jiang, Yitong and Zhang, Zhaoyang and Xue, Tianfan and Gu, Jinwei},
  booktitle={European Conference on Computer Vision},
  pages={340--359},
  year={2024},
  organization={Springer}
}

@inproceedings{Fei_2023_CVPR,
    author    = {Fei, Ben and Lyu, Zhaoyang and Pan, Liang and Zhang, Junzhe and Yang, Weidong and Luo, Tianyue and Zhang, Bo and Dai, Bo},
    title     = {Generative Diffusion Prior for Unified Image Restoration and Enhancement},
    booktitle = {Proceedings of the IEEE/CVF Conference on Computer Vision and Pattern Recognition (CVPR)},
    month     = {June},
    year      = {2023},
    pages     = {9935-9946}
}

@inproceedings{gou2024test,
  title={Test-Time Degradation Adaptation for Open-Set Image Restoration},
  author={Gou, Yuanbiao and Zhao, Haiyu and Li, Boyun and Xiao, Xinyan and Peng, Xi},
  booktitle={International Conference on Machine Learning},
  pages={16167--16177},
  year={2024}
}

@inproceedings{li2025ld,
  title={LD-RPS: Zero-Shot Unified Image Restoration via Latent Diffusion Recurrent Posterior Sampling},
  author={Huaqiu Li and Yong Wang and Tongwen Huang and Hailang Huang and Haoqian Wang and Xiangxiang Chu},
  booktitle = {Proceedings of the IEEE/CVF International Conference on Computer Vision (ICCV)},
  year={2025}
}

@inproceedings{Qin_2025_CVPR,
    author    = {Qin, Haina and Luo, Wenyang and Wang, Libin and Zheng, Dandan and Chen, Jingdong and Yang, Ming and Li, Bing and Hu, Weiming},
    title     = {Reversing Flow for Image Restoration},
    booktitle = {Proceedings of the IEEE/CVF Conference on Computer Vision and Pattern Recognition (CVPR)},
    month     = {June},
    year      = {2025},
    pages     = {7545-7558}
}

@misc{franzen1999kodak,
  author = {Rich Franzen},
  title = {Kodak Lossless True Color Image Suite},
  year = {1999},
  url = {\url{http://r0k.us/graphics/kodak/}}
}

@article{liu2023evaluating,
  title={Evaluating the generalization ability of super-resolution networks},
  author={Liu, Yihao and Zhao, Hengyuan and Gu, Jinjin and Qiao, Yu and Dong, Chao},
  journal={IEEE Transactions on pattern analysis and machine intelligence},
  volume={45},
  number={12},
  pages={14497--14513},
  year={2023},
  publisher={IEEE}
}

@inproceedings{wang2020deep,
  title={Deep degradation prior for low-quality image classification},
  author={Wang, Yang and Cao, Yang and Zha, Zheng-Jun and Zhang, Jing and Xiong, Zhiwei},
  booktitle={Proceedings of the IEEE/CVF Conference on Computer Vision and Pattern Recognition (CVPR)},
  pages={11049--11058},
  year={2020}
}

@article{wu2024ddr,
  title={Ddr: Exploiting deep degradation response as flexible image descriptor},
  author={Wu, Juncheng and Ni, Zhangkai and Wang, Hanli and Yang, Wenhan and Zhou, Yuyin and Wang, Shiqi},
  journal={Advances in Neural Information Processing Systems},
  volume={37},
  pages={61040--61064},
  year={2024}
}

@article{dhariwal2021diffusion,
  title={Diffusion models beat gans on image synthesis},
  author={Dhariwal, Prafulla and Nichol, Alexander},
  journal={Advances in neural information processing systems},
  volume={34},
  pages={8780--8794},
  year={2021}
}

@inproceedings{cao2024grids,
  title={Grids: Grouped multiple-degradation restoration with image degradation similarity},
  author={Cao, Shuo and Liu, Yihao and Zhang, Wenlong and Qiao, Yu and Dong, Chao},
  booktitle={European Conference on Computer Vision},
  pages={70--87},
  year={2024}
}

@inproceedings{agnolucci2024arniqa,
  title={ARNIQA: Learning Distortion Manifold for Image Quality Assessment},
  author={Agnolucci, Lorenzo and Galteri, Leonardo and Bertini, Marco and Del Bimbo, Alberto},
  booktitle={Proceedings of the IEEE/CVF Winter Conference on Applications of Computer Vision},
  pages={189--198},
  year={2024}
}

@inproceedings{johnson2016perceptual,
  title={Perceptual losses for real-time style transfer and super-resolution},
  author={Johnson, Justin and Alahi, Alexandre and Fei-Fei, Li},
  booktitle={European conference on computer vision},
  pages={694--711},
  year={2016}
}

@article{goodfellow2020generative,
  title={Generative adversarial networks},
  author={Goodfellow, Ian and Pouget-Abadie, Jean and Mirza, Mehdi and Xu, Bing and Warde-Farley, David and Ozair, Sherjil and Courville, Aaron and Bengio, Yoshua},
  journal={Communications of the ACM},
  volume={63},
  number={11},
  pages={139--144},
  year={2020},
  publisher={ACM New York, NY, USA}
}

@inproceedings{NEURIPS2019_bdbca288,
 author = {Paszke, Adam and Gross, Sam and Massa, Francisco and Lerer, Adam and Bradbury, James and Chanan, Gregory and Killeen, Trevor and Lin, Zeming and Gimelshein, Natalia and Antiga, Luca and Desmaison, Alban and Kopf, Andreas and Yang, Edward and DeVito, Zachary and Raison, Martin and Tejani, Alykhan and Chilamkurthy, Sasank and Steiner, Benoit and Fang, Lu and Bai, Junjie and Chintala, Soumith},
 booktitle = {Advances in Neural Information Processing Systems},
 pages = {},
 publisher = {Curran Associates, Inc.},
 title = {PyTorch: An Imperative Style, High-Performance Deep Learning Library},
 volume = {32},
 year = {2019}
}

@inproceedings{Rombach_2022_CVPR,
    author    = {Rombach, Robin and Blattmann, Andreas and Lorenz, Dominik and Esser, Patrick and Ommer, Bj\"orn},
    title     = {High-Resolution Image Synthesis With Latent Diffusion Models},
    booktitle = {Proceedings of the IEEE/CVF Conference on Computer Vision and Pattern Recognition (CVPR)},
    month     = {June},
    year      = {2022},
    pages     = {10684-10695}
}

@inproceedings{zhang2018unreasonable,
  title={The unreasonable effectiveness of deep features as a perceptual metric},
  author={Zhang, Richard and Isola, Phillip and Efros, Alexei A and Shechtman, Eli and Wang, Oliver},
  booktitle={Proceedings of the IEEE conference on computer vision and pattern recognition},
  pages={586--595},
  year={2018}
}

@article{wang2004image,
  title={Image quality assessment: from error visibility to structural similarity},
  author={Wang, Zhou and Bovik, Alan C and Sheikh, Hamid R and Simoncelli, Eero P},
  journal={IEEE transactions on image processing},
  volume={13},
  number={4},
  pages={600--612},
  year={2004},
  publisher={IEEE}
}

@inproceedings{Wang_2018_ECCV_Workshops,
author = {Wang, Xintao and Yu, Ke and Wu, Shixiang and Gu, Jinjin and Liu, Yihao and Dong, Chao and Qiao, Yu and Change Loy, Chen},
title = {ESRGAN: Enhanced Super-Resolution Generative Adversarial Networks},
booktitle = {Proceedings of the European Conference on Computer Vision (ECCV) Workshops},
month = {September},
year = {2018}
}

@article{mittal2012making,
  title={Making a “completely blind” image quality analyzer},
  author={Mittal, Anish and Soundararajan, Rajiv and Bovik, Alan C},
  journal={IEEE Signal processing letters},
  volume={20},
  number={3},
  pages={209--212},
  year={2012},
  publisher={IEEE}
}

@inproceedings{Li_2022_CVPR,
    author    = {Li, Boyun and Liu, Xiao and Hu, Peng and Wu, Zhongqin and Lv, Jiancheng and Peng, Xi},
    title     = {All-in-One Image Restoration for Unknown Corruption},
    booktitle = {Proceedings of the IEEE/CVF Conference on Computer Vision and Pattern Recognition (CVPR)},
    month     = {June},
    year      = {2022},
    pages     = {17452-17462}
}

@inproceedings{potlapalli2023promptir,
  title={PromptIR: Prompting for All-in-One Image Restoration},
  author={Potlapalli, Vaishnav and Zamir, Syed Waqas and Khan, Salman and Khan, Fahad},
  booktitle={Thirty-seventh Conference on Neural Information Processing Systems},
  year={2023}
}

@inproceedings{Zheng_2024_CVPR,
    author    = {Zheng, Dian and Wu, Xiao-Ming and Yang, Shuzhou and Zhang, Jian and Hu, Jian-Fang and Zheng, Wei-Shi},
    title     = {Selective Hourglass Mapping for Universal Image Restoration Based on Diffusion Model},
    booktitle = {Proceedings of the IEEE/CVF Conference on Computer Vision and Pattern Recognition (CVPR)},
    month     = {June},
    year      = {2024},
    pages     = {25445-25455}
}

@article{Li:2021kt,
  title={You only look yourself: Unsupervised and untrained single image dehazing neural network},
  author={Li, Boyun and Gou, Yuanbiao and Gu, Shuhang and Liu, Jerry Zitao and Zhou, Joey Tianyi and Peng, Xi},
  journal={International Journal of Computer Vision},
  volume={129},
  number={5},
  pages={1754--1767},
  year={2021},
  publisher={Springer}
}

@article{li2020zero,
  title={Zero-shot image dehazing},
  author={Li, Boyun and Gou, Yuanbiao and Liu, Jerry Zitao and Zhu, Hongyuan and Zhou, Joey Tianyi and Peng, Xi},
  journal={IEEE Transactions on Image Processing},
  volume={29},
  pages={8457--8466},
  year={2020},
  publisher={IEEE}
}

@inproceedings{Gandelsman_2019_CVPR,
author = {Gandelsman, Yosef and Shocher, Assaf and Irani, Michal},
title = {"Double-DIP": Unsupervised Image Decomposition via Coupled Deep-Image-Priors},
booktitle = {Proceedings of the IEEE/CVF Conference on Computer Vision and Pattern Recognition (CVPR)},
month = {June},
year = {2019}
}

@inproceedings{Shi_2024_CVPR,
    author    = {Shi, Yiqi and Liu, Duo and Zhang, Liguo and Tian, Ye and Xia, Xuezhi and Fu, Xiaojing},
    title     = {ZERO-IG: Zero-Shot Illumination-Guided Joint Denoising and Adaptive Enhancement for Low-Light Images},
    booktitle = {Proceedings of the IEEE/CVF Conference on Computer Vision and Pattern Recognition (CVPR)},
    month     = {June},
    year      = {2024},
    pages     = {3015-3024}
}

@inproceedings{Liang_2023_ICCV,
    author    = {Liang, Zhexin and Li, Chongyi and Zhou, Shangchen and Feng, Ruicheng and Loy, Chen Change},
    title     = {Iterative Prompt Learning for Unsupervised Backlit Image Enhancement},
    booktitle = {Proceedings of the IEEE/CVF International Conference on Computer Vision (ICCV)},
    month     = {October},
    year      = {2023},
    pages     = {8094-8103}
}

@article{li2021learning,
  title={Learning to enhance low-light image via zero-reference deep curve estimation},
  author={Li, Chongyi and Guo, Chunle and Loy, Chen Change},
  journal={IEEE transactions on pattern analysis and machine intelligence},
  volume={44},
  number={8},
  pages={4225--4238},
  year={2021},
  publisher={IEEE}
}

@article{li2018benchmarking,
  title={Benchmarking single-image dehazing and beyond},
  author={Li, Boyi and Ren, Wenqi and Fu, Dengpan and Tao, Dacheng and Feng, Dan and Zeng, Wenjun and Wang, Zhangyang},
  journal={IEEE transactions on image processing},
  volume={28},
  number={1},
  pages={492--505},
  year={2018},
  publisher={IEEE}
}

@inproceedings{Ma_2022_CVPR,
    author    = {Ma, Long and Ma, Tengyu and Liu, Risheng and Fan, Xin and Luo, Zhongxuan},
    title     = {Toward Fast, Flexible, and Robust Low-Light Image Enhancement},
    booktitle = {Proceedings of the IEEE/CVF Conference on Computer Vision and Pattern Recognition (CVPR)},
    month     = {June},
    year      = {2022},
    pages     = {5637-5646}
}

@inproceedings{Li_2017_ICCV,
author = {Li, Boyi and Peng, Xiulian and Wang, Zhangyang and Xu, Jizheng and Feng, Dan},
title = {AOD-Net: All-In-One Dehazing Network},
booktitle = {Proceedings of the IEEE International Conference on Computer Vision (ICCV)},
month = {Oct},
year = {2017}
}

@inproceedings{Lee_2022_CVPR,
    author    = {Lee, Wooseok and Son, Sanghyun and Lee, Kyoung Mu},
    title     = {AP-BSN: Self-Supervised Denoising for Real-World Images via Asymmetric PD and Blind-Spot Network},
    booktitle = {Proceedings of the IEEE/CVF Conference on Computer Vision and Pattern Recognition (CVPR)},
    month     = {June},
    year      = {2022},
    pages     = {17725-17734}
}

@inproceedings{Jang_2021_ICCV,
    author    = {Jang, Geonwoon and Lee, Wooseok and Son, Sanghyun and Lee, Kyoung Mu},
    title     = {C2N: Practical Generative Noise Modeling for Real-World Denoising},
    booktitle = {Proceedings of the IEEE/CVF International Conference on Computer Vision (ICCV)},
    month     = {October},
    year      = {2021},
    pages     = {2350-2359}
}

@inproceedings{jang2023puca,
  title={PUCA: Patch-Unshuffle and Channel Attention for Enhanced Self-Supervised Image Denoising},
  author={Jang, Hyemi and Park, Junsung and Jung, Dahuin and Lew, Jaihyun and Bae, Ho and Yoon, Sungroh},
  booktitle={Thirty-seventh Conference on Neural Information Processing Systems},
  year={2023}
}

@article{ma2024masked,
  title={Masked pre-training enables universal zero-shot denoiser},
  author={Ma, Xiaoxiao and Wei, Zhixiang and Jin, Yi and Ling, Pengyang and Liu, Tianle and Wang, Ben and Dai, Junkang and Chen, Huaian},
  journal={Advances in Neural Information Processing Systems},
  volume={37},
  pages={32570--32610},
  year={2024}
}

@InProceedings{Yang_2025_ICCV,
    author    = {Yang, Jing and Xing, Qunliang and Xu, Mai and Qiao, Minglang},
    title     = {Uncover Treasures in DCT: Advancing JPEG Quality Enhancement by Exploiting Latent Correlations},
    booktitle = {Proceedings of the IEEE/CVF International Conference on Computer Vision (ICCV)},
    month     = {October},
    year      = {2025},
    pages     = {17598-17607}
}
